\documentclass[Afour, sageh, times]{sagej}
\usepackage{moreverb,url}
\usepackage{subcaption}
\usepackage[colorlinks,bookmarksopen,bookmarksnumbered,citecolor=black,urlcolor=black]{hyperref}
\usepackage{enumerate}
\usepackage{color}
\usepackage{multirow}

\usepackage{array}
\newcolumntype{L}[1]{>{\raggedright\let\newline\\\arraybackslash\hspace{0pt}}m{#1}}
\newcolumntype{C}[1]{>{\centering\let\newline\\\arraybackslash\hspace{0pt}}m{#1}}
\newcolumntype{R}[1]{>{\raggedleft\let\newline\\\arraybackslash\hspace{0pt}}m{#1}}

\newcommand\BibTeX{{\rmfamily B\kern-.05em \textsc{i\kern-.025em b}\kern-.08ems
T\kern-.1667em\lower.7ex\hbox{E}\kern-.125emX}}

\newcommand{\nostarnote}[1]{}
\newcommand{\nonote}[1]{}
\def\volumenumber{Pri-print}
\def\issuenumber{2019}
\def\journalname{To apear at the IJRR.} 

\setcitestyle{aysep={,}} % comma between author, year  
\begin{document}

\runninghead{Islam et al.}

\title{Person Following by Autonomous Robots: A Categorical Overview}

\author{Md Jahidul Islam, Jungseok Hong and Junaed Sattar}

\affiliation{The authors are with the Interactive Robotics and Vision Laboratory (IRVLab),  University of Minnesota, Twin Cities, MN, USA.}

\corrauth{Md Jahidul Islam (Email: islam034@umn.edu)}

%\email{islam034@umn.edu}

\begin{abstract}
A wide range of human-robot collaborative applications in diverse domains such as manufacturing, health care, the entertainment industry, and social interactions, require an autonomous robot to follow its human companion. 
Different working environments and applications pose diverse challenges by adding constraints on the choice of sensors, the degree of autonomy, and dynamics of a person-following robot. Researchers have addressed these challenges in many ways and contributed to the development of a large body of literature. This paper provides a comprehensive overview of the literature by categorizing different aspects of person-following by autonomous robots. Also, the corresponding operational challenges are identified based on various design choices for ground, underwater, and aerial scenarios. In addition, state-of-the-art methods for perception, planning, control, and interaction are elaborately discussed and their applicability in varied operational scenarios are presented. Then, some of the prominent methods are qualitatively compared, corresponding practicalities are illustrated, and their feasibility is analyzed for various use-cases. Furthermore, several prospective application areas are identified, and open problems are highlighted for future research.
\end{abstract}


\maketitle

\section{Introduction}\label{sec:intro}
\emph{Person following} scenarios arise when a human and an autonomous robot collaborate on a common task that requires the robot to follow the human. Usually, the human leads the task and cooperates with the robot during task execution. An example application would be the service robots, which are widely used in industrial applications, \emph{e.g.}, in manufacturing, warehousing, and health care. The use of companion robots in surveillance, social interaction, and medical applications has also flourished over the last decade. Numerous new applications are also emerging in the entertainment industry as robots are getting more accessible for personal use. 

Based on the operating medium, person-following can be categorized into ground, underwater, and aerial scenarios. Ground service robots following a human while performing a cooperative task is the canonical example of person-following. Such assistant robots are being used in many domestic and industrial applications~\citep{piaggio},
and in health care \citep{ilias2014nurse,iribe2011study}. Moreover, diver-following robots are useful in the submarine pipeline and shipwreck inspection, marine life and seabed monitoring, and many other underwater exploration activities~\citep{islam2018towards,sattar2009underwater,miskovic2015tracking}. Furthermore, the use of person-following aerial robots ~\citep{mueller2016persistent,pestana2014computer,naseer2013followme} has flourished over the last decade in the entertainment industry~\citep{Skydio} as quadcopters have become quite popular for filming outdoor activities such as mountain climbing, biking, surfing, and many other sporting endeavors.

In all applications mentioned above, following a person is not the primary objective of the robot, yet it is vital for achieving the overall success of the primary operation. Robust techniques to enable person-following are thus of significant importance in the repertoire of robotic behaviors. The major computational components of a person-following system are perception, planning, control, and interaction. The design of each of these components largely depends on the choice of sensors and the degree of autonomy for the robot. Additionally, different scenarios (\textit{i.e.}, ground, underwater, and aerial) pose different operational challenges and add constraints on the construction and dynamics of the robot. These factors are determined by specific application requirements which make it difficult to design a generic person-following methodology. 

Attempts to develop person-following robots for a wide range of applications have resulted in a variety of different methodologies. 
Particularly, the computer vision and robotics communities have been developing person-following methodologies for ground robots since the nineties~\citep{darrell1998integrated,wren1997pfinder,azarbayejani1996real}. Initially seen as a special case of object tracking, person-following by autonomous robots soon became a challenging problem of its own as many industrial applications started to flourish~\citep{pairo2013person,cu2013human,ess2008mobile,balan2005quantitative}. Recently, other aspects of the problem such as human-robot interaction, social awareness, the degree of autonomy, etc., are getting attention from the research community as well~\citep{triebel2016spencer,dewantara2016generation}. The advent of underwater and aerial applications has added other dimensions to this growing field \citep{mueller2016persistent,naseer2013followme, sattar2009underwater}. Different mediums and a diverse set of operational considerations often demand application-specific design for a person-following robot. However, certain design issues, underlying algorithms, methods of human-robot interaction, among others, remain mostly generic for all person-following scenarios. An elaborate discussion on these aspects, comparison of different approaches and the state-of-the-art techniques would greatly help the current and future researchers.

This paper outlines various aspects of the person-following problem and provides a comprehensive overview of the existing literature. In addition, different issues pertaining to the robot and algorithmic design are identified, operational scenarios are illustrated, and qualitative analysis of the state-of-the-art approaches are presented. Specifically, the contributions of this paper are the following:

\begin{itemize}
\item A categorization of the person-following problem is presented based on various attributes such as the medium of operation, choice of sensors, mode of interaction, the degree of autonomy, etc. Operational scenarios for each category are then discussed along with the relevant applications.

\item Additionally, for different person-following scenarios, key design issues are identified, the underlying assumptions are discussed, and state-of-the-art approaches to cope with the operational challenges are presented. 

\item Subsequently, an elaborate discussion on the underlying algorithms of different state-of-the-art approaches for perception, planning, control, and interaction are presented. 
The attributes and overall feasibility of these algorithms are qualitatively analyzed and then compared based on various operational considerations.

\item Furthermore, several open problems for future research are highlighted along with their current status in the literature. 
\end{itemize}
\section{Categorization of Autonomous Person-Following Behaviors}\label{sec:cate}
Person-following behaviors by autonomous robots can be diverse depending on several application-specific factors such as the medium of operation, choice of sensors, mode of interaction,  granularity, and the degree of autonomy. 
The design and overall operation of a person-following robot mostly depend on the operating medium, \textit{e.g.}, ground, underwater, and aerial. Other application-specific constraints influence the choice of sensors, mode of interaction (explicit or implicit), granularity, and degree of autonomy (full or partial). In this paper, \textit{explicit} and \textit{implicit} interactions refer to direct and indirect human-robot communication, respectively. In addition, the term \textit{granularity} is referred to as the number of humans and robots involved in a person-following scenario.

\begin{figure*}[ht]
\centering
\includegraphics [width=\linewidth]{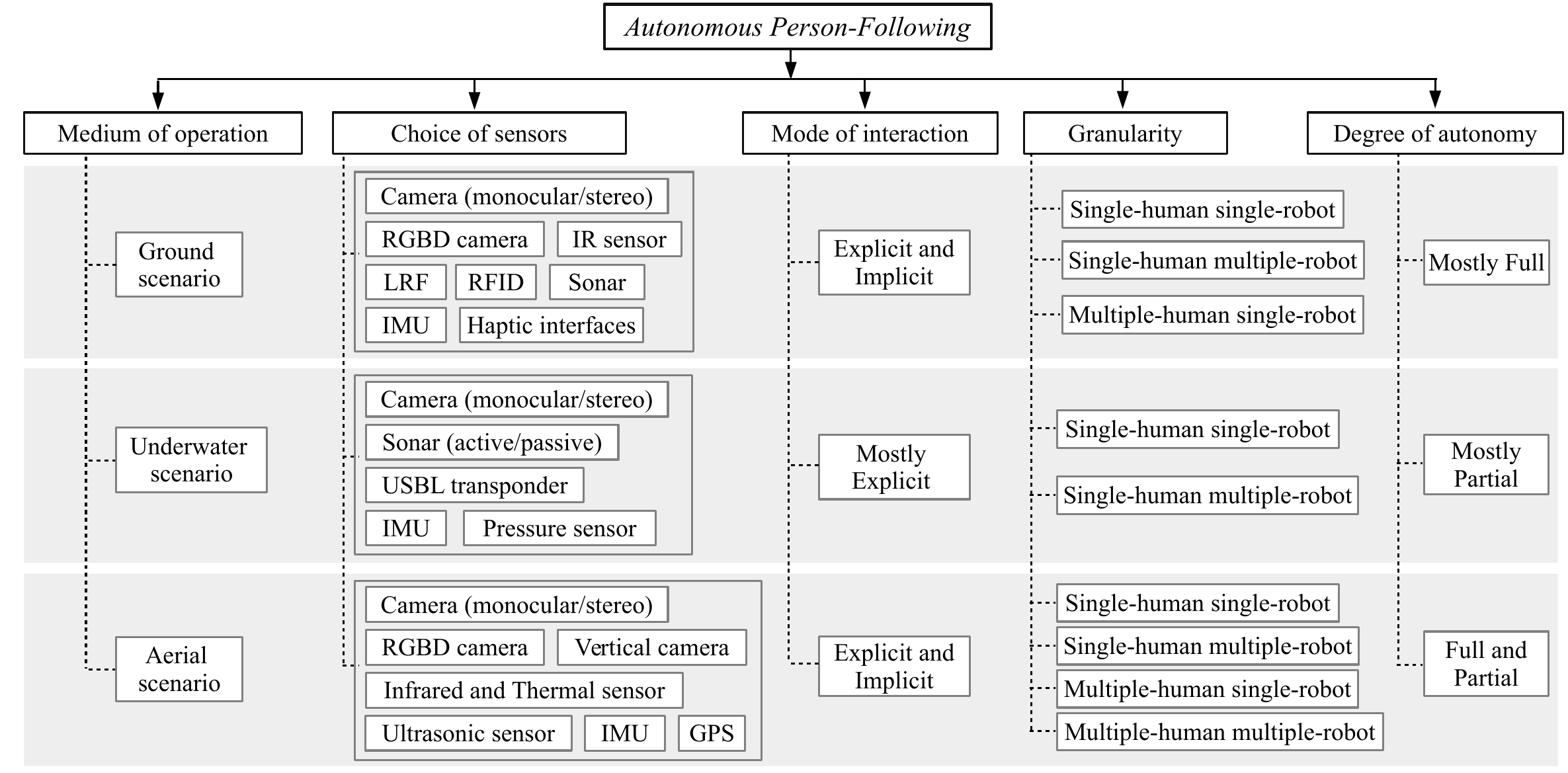}
\vspace{-6mm}
\caption{A categorization of autonomous person-following behaviors based on five major attributes: the medium of operation, choice or sensors, mode of interaction, granularity, and degree of autonomy. }
\label{fig:cat}
\end{figure*}

Based on the above-mentioned attributes, a simplified categorization of autonomous person-following behaviors is depicted in Figure~\ref{fig:cat}. The rest of this paper is organized by following the categorization based on the medium of operation, while other attributes are discussed as subcategories.

\subsection{Ground Scenario}
\begin{figure}[ht]
    \centering
    \begin{subfigure}[t]{0.184\textwidth}
        \includegraphics[width=\linewidth]{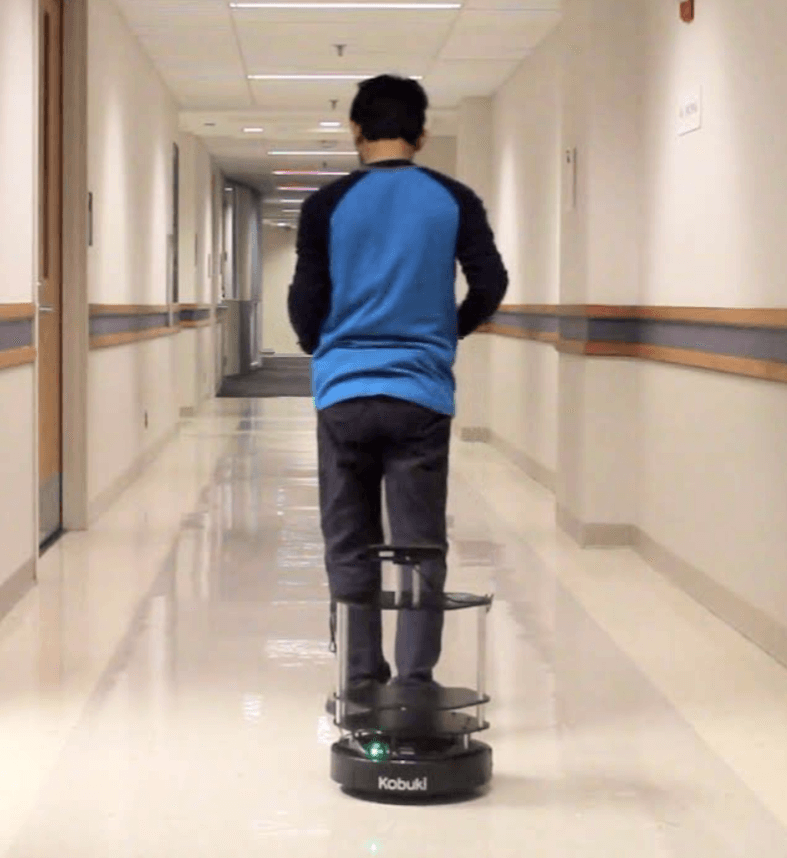}
        \caption{Indoor}
    \end{subfigure}%
    ~
    \begin{subfigure}[t]{0.302\textwidth}
        \centering
        \includegraphics[width=\linewidth]{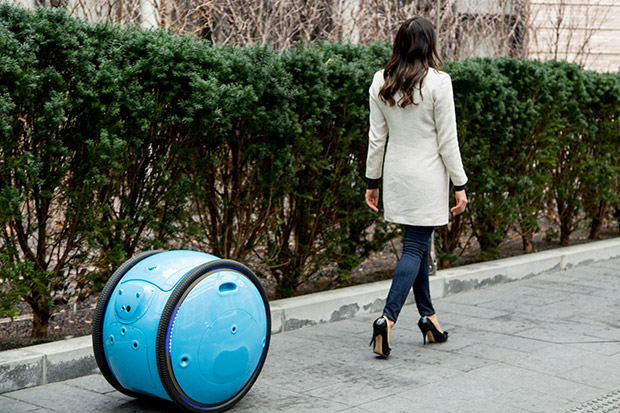}
        \caption{Outdoor}
    \end{subfigure}
    \vspace{-2mm}
    \caption{Typical person-following scenarios for ground robots: (a) TurtleBot \citep{turtle} is following a person in an indoor setting; (b) Gita cargo robot \citep{piaggio} is following a person outdoors.}
    \label{fig:ground}
\end{figure}

Domestic assistant robots~\citep{piaggio} and shopping-cart robots~\citep{nishimura2007development} are the most common examples of person-following UGVs (Unmanned Ground Vehicles). Their usage in several other industrial applications~\citep{budgee}, and in health care is also increasing in recent times~\citep{tasaki2015prototype,ilias2014nurse,iribe2011study}. Figure~\ref{fig:ground} illustrates typical person-following scenarios for ground robots. The UGV uses its camera and other sensors to detect the person in its field-of-view. Once the position and distance of the person are approximated, path planning is performed, and the corresponding control signals are generated in order to follow the person. Details on these operations and the related state-of-the-art methodologies will be discussed later in this paper. The following discussion expounds various design issues and related operational constraints based on the choice of sensors, mode of interaction, granularity, and degree of autonomy.

\subsubsection{Choice of Sensors} 
Most person-following UGVs are equipped with cameras and the perception is performed via visual sensing. Other sensors are used to accurately measure the distance and activities (walking, waving hands, etc.) of the person for safe navigation and interaction. Choice of sensors is often determined by the operating environment, \textit{i.e.}, indoor or outdoor. For example, RGBD sensors are very effective in indoor environments; in addition to having a regular RGB camera, they are equipped with an IR sensor to provide the associated depth information~\citep{mi2016system}. Therefore, both the position and distance of the person can be measured with high accuracy. However, since IR sensors perform poorly in the presence of sunlight, they are not good choices for outdoor environments. The use of stereo cameras can get rid of this problem as depth information can be approximated by using the triangulation techniques~\citep{chen2017person,satake2013visual}. Laser Range Finders (LRFs) are also widely used by person-following UGVs~\citep{susperregi2013rgb,chung2012detection}. These sensors provide a cluster of directional distance measures, from which the presence and distance of a person can be approximated. Moreover, some applications use Ultra-Wide Band (UWB)~\citep{laneurit2016trackbod}, RFID tags~\citep{germa2010vision,kulykukin2004human}, and haptic interfaces~\citep{ghosh2014following} as exteroceptive sensors. 

Proprioceptive sensors such as Inertial Measurement Units (IMUs) are used to keep track of the robot's relative motion and orientation information~\citep{brookshire2010person} for smooth and accurate navigation. Researchers have also explored the use of wearable IMUs for modeling the human walking gait~\citep{cifuentes2014human}, which is useful for differentiating humans from other moving objects. 
 
Person-following UGVs typically use multiple sensors in order to ensure robustness and efficiency. Standard sensor fusion techniques are then adopted to reduce the uncertainty in sensing and estimation~\citep{cifuentes2014human,susperregi2013rgb,luo2009human}. A summary of the key design issues based on different choices of sensors for person-following UGVs is presented in Table~\ref{GroundSensors}.

\subsubsection{Mode of Interaction}
It is ideal for a person-following robot to interact with the human user in a natural way. Even if the robot is not designed to interact with the person directly, implicit (\textit{i.e.}, indirect) interactions exist since the human is aware that a robot is following them~\citep{hu2014design}. This awareness is important for the overall success of the operation because the person can simplify the robot's task in many ways, \textit{e.g.}, by making smooth turns, avoiding obstacles, walking with reasonable speed, etc. Additionally, the robot needs to maintain a safe distance and plan a socially-aware motion trajectory while following the person~\citep{triebel2016spencer,granata2012framework}. A number of user studies have explored different aspects of implicit interactions including but not limited to the robot's spatial conduct~\citep{fleishman2018proxemic}, preferred following angles~\citep{shanee2016influence}, turning behaviors~\citep{hu2014design}, socially-aware motion conduct~\citep{honig2018towards,triebel2016spencer}, etc. An elaborate discussion about these aspects is provided later in this paper (see Section~\ref{implicitI}).

Explicit interactions refer to direct human-robot communication. In many practical applications, a human instructs the UGV to perform certain tasks such as changing its motion or speed, taking photographs, making phone calls, etc. These instructions are typically communicated using voice commands~\citep{fritsch2004audiovisual}, hand gestures~\citep{marge2011comparing,doisy2013spatially}, or haptic interfaces~\citep{ghosh2014following,park2010towards}. 
Moreover, some smart carts and autonomous luggage robots allow users to interact using smartphone applications~\citep{digiacomcantonio2014self}.
Explicit human-robot interaction is essential for most person-following ground applications; however, it requires additional computational capabilities in order for the UGVs to understand human instructions and spontaneously interact in a natural manner. Table~\ref{GroundInteraction} highlights the challenges and responsibilities involved in different forms of human-robot interactions for person-following UGVs.

\begin{table*}[ht]
\centering
\footnotesize
\caption{Choice of sensors and the corresponding design issues for person-following UGVs.}
\begin{tabular}{|p{1.6cm}||p{1.9cm}|p{2.6cm}|p{4.4cm}|p{4.5cm}|} % total 15
\hline
\textbf{Sensor}  & \textbf{Data} & \textbf{Challenges/Limitations}  & \textbf{Usage/Operation} & \textbf{{   Selected References }} \\
\hline \hline
Monocular camera & RGB image & Low visibility; lighting variation  & Computer vision-based algorithms are used for detection and tracking  & \cite{pierre2018end,guevara2016vision,isobe2014human,ma2008sensor,kobilarov2006people,kwon2005person} \\ 
\hline
Stereo camera  & RGB image & Low visibility; lighting variation  & In addition to RGB image-based detection and tracking, stereo triangulation techniques are used to approximate the associated depth information  &  \cite{chen2017person,satake2013visual,hu2014design,satake2012sift,brookshire2010person,satake2009robust,luo2009human,takemura2007person,itoh2006development} \\ 
\hline
RGBD camera &  RGBD data & Presence of sunlight  & In addition to RGB image-based detection and tracking, distance of the person is approximated using the depth data   & \cite{wang2017real,masuzawa2017development,mi2016system,basso2013fast,cosgun2013autonomous,munaro2013software,doisy2012adaptive} \\ 
\hline
LRF  & Planner distance measures  & Presence of transparent (\textit{e.g.}, glass) or dark surfaces   & Person's body is detected from a cluster of distance measures   & \cite{leigh2015person,cai2014human,cosgun2013autonomous,pairo2013person,jung2012control,alvarez2012feature,shaker2008fuzzy} \\ 
\hline
Sonar & Directional distance measures &  Specular reflections; crosstalk  & Presence and distance of a person is detected from the directional distance measures  & \cite{peng2016tracking,gascuena2011agent,itoh2006development}  \\ 
\hline
RFID & RF signal &  Presence of interfering signals; limited range; ensuring privacy    &  Person carrying an RFID tag is tracked by following the direction of the RFID signal source  & \cite{germa2010vision,kulykukin2004human} \\ 
\hline
IMU & IMU data &  Precision; drift  &  Robot's relative orientation, angular and linear velocity, and acceleration are estimated for motion control    & \cite{brookshire2010person,cifuentes2014human} \\ 
\hline
\end{tabular}
\label{GroundSensors}
\end{table*}

\begin{table*}[ht]
\centering
\footnotesize
\caption{Challenges and responsibilities involved in implicit and explicit human-robot interactions for person-following UGVs.}
\begin{tabular}{|p{1.4cm}||p{4.0cm}|p{4.0cm}|p{6.0cm}|}
\hline
\textbf{Interaction}  & \textbf{Challenges for the Robot} & \textbf{Responsibilities of the Person}  & \textbf{{   Selected References}} \\
\hline \hline
Implicit  & Maintaining safe distance and speed; socially-aware spatial and motion conduct   planning & Walking with reasonable speed; avoiding obstacles; taking smooth turns  & \cite{honig2018towards,triebel2016spencer,ferrer2013robot,liem2008hybrid,gockley2007natural,kuno2007museum,matsumaru2005mobile,kulykukin2004human} \\ 
\hline
Explicit  & \textit{Additional challenges:} recognizing and decoding instruction commands; planning and acting based on given instructions   &  \textit{Additional responsibility:} communicating clearly based on the predefined scheme  & \cite{triebel2016spencer,ghosh2014following,doisy2013spatially,cosgun2013autonomous,marge2011comparing,hirai2003visual} \\ 
\hline
\end{tabular}
\label{GroundInteraction}
\end{table*}

\subsubsection{Granularity}
Most domestic applications require a single robot accompanying a single person. Interacting with a specific person is common as well, particularly for accompanying the elderly and people with disabilities~\citep{ilias2014nurse,liem2008hybrid,kulykukin2004human}. The most important features of these robots are the social and interactive skills needed to make the accompanying person feel safe and attended to. In industrial applications, however, robustness and performance are more important relative to their social aspects~\citep{cosgun2013autonomous,germa2010vision}. These robots typically assist a single person in a dynamic multi-agent environment, \textit{i.e.}, with the presence of other humans and robots.

A robot can also accompany a group of people by following the \textit{center of attention} of the group~\citep{chen2017person,basso2013fast}. However, this can be challenging if people move in random directions. An anchorperson is generally specified to the robot beforehand who interacts with the robot and helps it to navigate. In such cases, the robot uses anchorperson-specific features for tracking while interacting with the group as a whole. 
Since interacting with a group of people can be challenging, service robots are often equipped with user interfaces for easy and effective human-robot interaction. 
Furthermore, multiple independent robots can assist a single person in a common task, given that the person is responsible for synchronizing their activities. Although swarm-like multi-robot cooperation or non-cooperative multi-agent synchronization~\citep{chen2017decentralized} is possible, these frameworks are rather resource-demanding and not commonly adopted in person-following applications.

\subsubsection{Degree of Autonomy}
One major advantage of using person-following robots is that it eliminates the need for dedicated teleoperation. Since autonomous ground navigation is relatively less challenging compared to underwater or aerial scenarios, person-following UGVs are typically designed to have fully autonomous behavior~\citep{leigh2015person}. 
Some applications, however, allow partial autonomy for UGVs that perform very specific tasks such as assisting a nurse in their operating room~\citep{ilias2014nurse}, serving food at a restaurant~\citep{pieska2013social}, etc. These service robots follow their companion around within a predefined operating area and provide assistance by carrying or organizing equipment, serving food, etc. 
While doing so, they may take human inputs for making navigation decisions such as when to follow, on which side to stay, where or when to wait, which objects to carry/organize, etc.

Such semi-autonomous behaviors for UGVs are adopted in robot-guiding applications as well, \textit{e.g.}, guiding a visually impaired person~\citep{ghosh2014following}, tour-guiding at a museums or a shopping mall~\citep{kanda2009affective,burgard1998interactive}, etc. Although robot-guiding is not strictly a person-following application, it shares a similar set of features and operational challenges for assisting a human companion. In particular, features such as socially aware planning, some aspects of explicit interaction, navigating through crowds while guiding/leading people, etc., are closely related to the person-following applications. The readers are referred to Table~\ref{all_sum} in advance for an organized and annotated collection of the person-following (and relevant) literature.

\subsection{Underwater Scenario}\label{sec:UW_scen}
Underwater missions are often conducted by a team of human divers and autonomous robots who cooperatively perform a set of common tasks~\citep{islam2018understanding,sattar2008enabling}. The divers typically lead the tasks and interact with the robots which follow the divers at certain stages of the mission~\citep{islam2018towards}. These situations arise in important applications such as the inspection of ship-hulls and submarine pipelines, the study of marine species migration, search-and-rescue, surveillance, etc. 
In these applications, following and interacting with the companion diver ~\citep{islam2018understanding} is essential because fully autonomous navigation is challenging due to the lack of radio communication and global positioning information underwater. Additionally, the human-in-the-loop guidance reduces operational overhead by eliminating the necessity of teleoperation or complex mission planning \textit{a priori}.

\begin{figure}[h]
    \centering
        \begin{subfigure}[t]{0.245\textwidth}
        \centering
        \includegraphics[width=\linewidth]{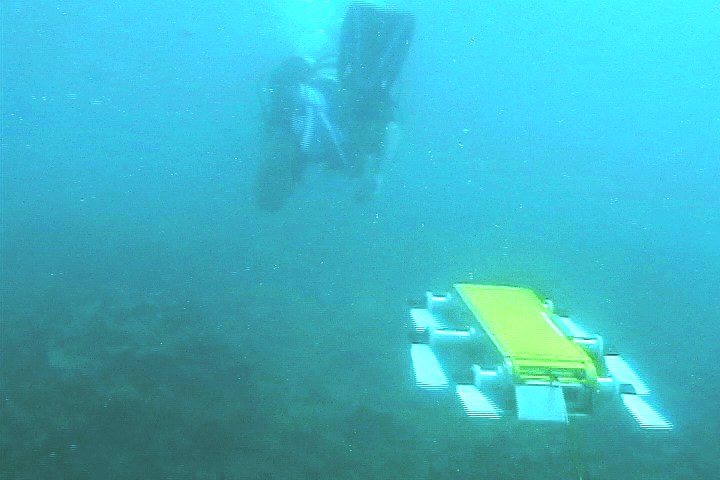}
        \caption{An underwater robot is following a diver}
    \end{subfigure}
    \begin{subfigure}[t]{0.225\textwidth}
        \includegraphics[width=\linewidth]{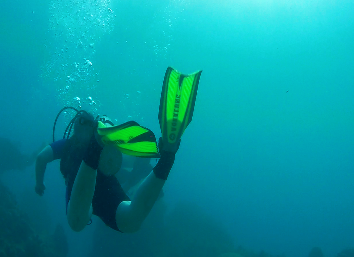}
        \caption{A diver seen from the robot's camera}
    \end{subfigure}
    \vspace{-2mm}
    \caption{A typical diver-following scenario for an underwater robot during a reef exploration task.}
    \label{fig:IntroPicture}
\end{figure}

Figure~\ref{fig:IntroPicture} illustrates a scenario where an Autonomous Underwater Vehicle (AUV) is following a scuba diver during an underwater mission. 
The operational complexities and risks involved in underwater applications are generally much higher compared to the ground applications~\citep{sattar2006performance}. The following Sections discuss these operational challenges and the related design issues based on the choice of sensors, mode of interaction, granularity, and degree of autonomy.

\begin{table*}[ht]
\centering
\footnotesize
\caption{Choice of sensors and the corresponding design issues for person-following underwater robots.}
\begin{tabular}{|p{1.7cm}||p{1.9cm}|p{3.0cm}|p{4.8cm}|p{3.6cm}|} %15
\hline
\textbf{Sensor}  & \textbf{Data} & \textbf{Challenges/Limitations}  & \textbf{Usage/Operation} & \textbf{{   Selected References}} \\
\hline \hline
Monocular camera & RGB image & Poor lighting conditions and visibility; suspended particles; color distortions  & Computer vision-based algorithms are used for detection and tracking  & \cite{islam2018towards,islam2017mixed,sattar2006performance}  \\ 
\hline
Stereo camera  & RGB image & Poor lighting conditions and visibility; suspended particles; color distortions  & In addition to RGB image-space based detection and tracking, stereo triangulation techniques are used to approximate the associated depth & \cite{stilinovic2015auv,corke2007experiments}  \\ 
\hline
Active sonar & Directional distance measures &  Noisy reading; scattering and reverberation  & Diver's presence and distance is approximated from the directional distance measures   & \cite{mandic2016underwater,demarco2013sonar} \\ 
\hline
Passive sonar & Frequency responses &  Extremely noisy reading; limited coverage  &  Frequency responses of the sound-wave generated by the diver are used for detection  &  \cite{hari2015robust,gemba2014partial} \\ 
\hline
USBL transponder & Acoustic pulse &  Presence of a paired USBL transceiver; noisy reading  &  Diver's position information is estimated by communicating with the transceiver   &  \cite{mandic2016underwater,miskovic2015tracking,corke2007experiments} \\ 
\hline
IMU & IMU data &  Precision; drift    &  Robot's relative orientation, angular and linear velocity, and acceleration are estimated for motion control    &  \cite{miskovic2015tracking} \\ 
\hline
Pressure sensor & Depth measurement &  Sensitiveness to temperature    &  Depth of the robot is approximated using the measured external pressure  &  \cite{corke2007experiments} \\ 
\hline
\end{tabular}
\label{UnderwaterSensors}
\end{table*}

\subsubsection{Choice of sensors}
Underwater diver-following robots usually rely on vision for tracking due to the bandwidth limitations of acoustic modems. In addition, it is undesirable to be intrusive and disruptive to the ecosystem~\citep{slabbekoorn2010noisy}. Cameras, being passive sensors (\textit{i.e.}, do not emit energy), are thus preferred over active sensors. Additionally, the use of stereo cameras is effective for approximating the relative distance of a diver or other targets~\citep{stilinovic2015auv,corke2007experiments}; standard computer vision-based techniques are then utilized for visual tracking. Although visibility can be a challenge, there is usually ample natural daylight at depths (typically $20$-$25$ meters) where humans can dive and remain for extended periods of time without using specialized equipment.

However, visual data gets noisy due to challenging marine conditions~\citep{sattar2006performance} arising from factors such as color distortions, lighting variations, suspended particles, etc. Consequently, robust visual detection and tracking become extremely difficult. Passive sonars such as hydrophones are useful in such scenarios~\citep{hari2015robust, gemba2014partial}. Active sonars are also used for diver-following in unfavorable visual conditions~\citep{mandic2016underwater, demarco2013sonar}. They are particularly useful when the robot loses the diver from its field-of-view and tries to rediscover them; once rediscovered, it can switch back to visual tracking. On the other hand, Ultra-short baseline (USBL) is often used for global positioning of underwater robots and ROVs (Remotely Operated Vehicles). A USBL transponder (mounted on the robot) communicates with a USBL transceiver (mounted on a pole under a ship or a boat) using acoustic signals. Phase-differencing methods are then used by the USBL to calculate positioning information (\textit{e.g.}, range, angle, etc.). The robot uses this information for navigation and tracking divers or other objects of interest.

Proprioceptive sensors such as IMUs are used by underwater robots as well for internal state estimation~\citep{miskovic2015tracking}; 
in addition, pressure sensors are used for measuring the depth of the robot (from surface) using external pressure~\citep{corke2007experiments}.
The depth information is useful for the depth-control and altitude-control modules of the robot. Moreover, inertial navigation systems (INSs) and other navigation sensors can be used to determine the robot's instantaneous pose and velocity information; however, these systems drift, thus require repeated correction using secondary sensing system(s). Table~\ref{UnderwaterSensors} summarizes the challenges and operational issues based on different choices of sensors for person-following underwater robots.

\subsubsection{Mode of interaction}\label{sec:UW_interaction}
Since truly autonomous underwater navigation is still an open challenge, explicit interaction with the accompanying diver is crucial for diver-following robots. 
Particularly in complex missions such as surveillance and rescue operations, robots can dynamically adjust their mission parameters by regularly communicating with the diver. In addition, some underwater exploration and data collection processes require close human supervision. 
In these scenarios, the divers typically instruct the robot to perform certain tasks (\textit{e.g.}, record snapshots, take samples, etc.) in different situations~\citep{islam2018understanding}. Although such communication paradigms are fairly straightforward in terrestrial settings, these are rather complex undertakings for underwater robots.

A number of communication frameworks have been proposed for underwater human-robot interaction. In RoboChat~\citep{dudek2007visual}, the divers use a set of AR-Tag markers to display a predefined sequence of symbolic patterns to the robot; these patterns are then mapped to a set of grammar rules defined for the language. A major limitation of such marker-based frameworks is that the markers need to be carried along and used in the correct order to produce instruction commands for the robot. Although the number of required markers can be reduced by incorporating additional shapes or motion signs with the marker~\citep{xu2008natural,sattar2007fourier}, this framework still involves a significant cognitive load on the diver. A simpler alternative is to use hand gestures to communicate with the robot~\citep{chiarella2018novel,islam2017dynamic}. It comes more naturally to divers because they already communicate with each other using hand gestures. On the other hand, robots can communicate emergency messages (\textit{e. g.}, low battery) and periodic updates to the diver using an on-board screen, flashing lights, etc.

The social and behavioral aspects of underwater missions are limited~\citep{wu2015towards}. However, implicit diver-robot interactions are vital for ensuring robot's safety and the overall success of the operation. The associated cognitive load on the divers is another important consideration for designing an interaction framework~\citep{islam2017dynamic,chiarella2015gesture}.

\subsubsection{Granularity}
As mentioned, the applications envisioned for underwater diver-following robots usually require a team of divers and robots. In most cases, each robot is assigned to one leader (usually a diver) who guides the robot during a mission~\citep{islam2018towards}. The leader can be another robot as well. For instance, a robot can follow another robot which is following a diver; such operations are referred to as robot convoying~\citep{shkurti2017underwater}. Robot convoying is useful when there are more robots than divers. Additionally, it is often more convenient and safer than multiple robots following a single diver. Underwater robots are usually not assigned to follow multiple divers because it requires complex motion planning; also, interacting with multiple humans simultaneously can be computationally demanding and often problematic.  

\subsubsection{Degree of autonomy}
Since underwater missions are strictly constrained by time and physical resources, most diver-following applications use semi-autonomous robots that take human inputs for making navigation decisions when needed. This reduces the overhead associated with underwater robot deployment and simplifies the associated mission planning. 
For simple applications, diver-following robots are typically programmed to perform only some basic tasks autonomously, \textit{e.g.}, following the diver, hovering, taking snapshots, etc. These programs and associated parameters are numbered and made known to the robot (and diver) beforehand. The diver leads the mission and instructs the robot to execute (one of) these programs during operation. For instance, a diver might instruct the robot to follow them to the operation zone, then to hover at a particular location of interest, take pictures, and eventually follow them back at the end of the mission. This interactive operational loop is very useful for simple applications such as exploration and data collection~\citep{islam2018understanding}. However, more autonomous capabilities are required for complex applications such as surveillance, monitoring the migration of marine species, etc. ROVs are typically deployed for these critical applications; these are connected to a surface vehicle (usually a ship or a boat) via an umbilical link that houses communications cables, an energy source, and enables power and information transfer.

\subsection{Aerial Scenario}
Unmanned Aerial Vehicles (UAVs) are traditionally used for surveillance, industrial, and military applications. More recently, UAVs have become more accessible and popular for entertainment purposes and in the film industry. 
They are very useful for capturing sports activities such as climbing or skiing from a whole new perspective~\citep{Skydio,Staaker,higuchi2011flying} without the need for teleoperation or a full-scale manned aerial vehicle. Another interesting application is to use person-following UAVs to provide external visual imagery, which allows athletes to gain a better understanding of their motions~\citep{higuchi2011flying}. These popular use-cases have influenced significant endeavor in research and development for affordable UAVs, and they have been at the forefront of person-following aerial drone industry in recent times.

\begin{figure}[ht]
    \centering
     \includegraphics[width=\linewidth]{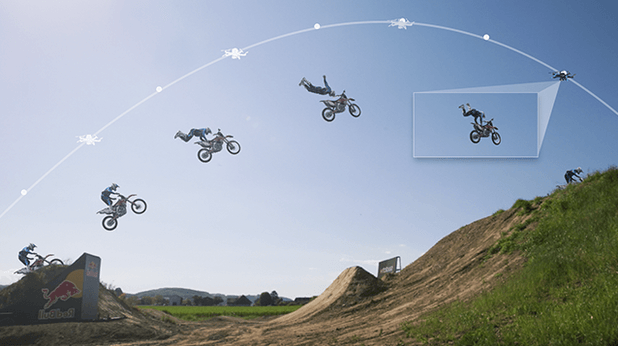}
      \vspace{-5mm}
        \caption{A UAV is filming a sport activity while intelligently following an athlete \citep{phantom}.} 
    \label{fig:AerialPicture}
\end{figure}

Figure \ref{fig:AerialPicture} illustrates a typical person-following scenario for UAVs. The operating time for UAVs is usually much shorter compared to the ground and underwater scenarios, \textit{e.g.}, less than half an hour to a few hours per episode due to limited battery life. The person launches the take-off command at the beginning of each episode and then commands it to follow (and possibly to capture snapshots) while he/she is performing some activities. The person makes the landing command after a reasonable amount of time and that ends the episode. It is common to carry multiple portable batteries or quick-chargers to capture longer events. The following sections discuss other operational considerations and related design issues based on the choice of sensors, mode of interaction, granularity, and degree of autonomy.

\subsubsection{Choice of sensors}
As the applications mentioned above suggest, person-following UAVs are equipped with cameras for visual sensing. Usually, a front-facing camera is used for this purpose, while an additional low-resolution vertical camera (\textit{i.e.}, facing down) is used as an optical-flow sensor. The vertical camera uses ground textures and visible features to determine the drone's ground velocity for ensuring stabilization. Due to the constraints on cost, weight, size, and battery life, use of other exteroceptive sensors are often limited to consumer-grade person-following UAVs. The Parrot ARDrone 2.0 \citep{ARDrone} for instance, only uses cameras (front and vertical) as exteroceptive sensors; these UAVs weigh less than a pound and cost approximately two hundred US dollars. On the other hand, with a 4K resolution camera and a three-axis mechanical gimbal, the DJI Mavic drones~\citep{Mavic} weigh $1.6$-$1.8$ pounds and cost approximately a thousand US dollars.

However, UAVs used in industrial, military, and other critical applications can accommodate multiple high-resolution cameras, range sensors, stereo cameras, etc. For instance, Inspire 2.0~\citep{Inspire} drones have additional upward facing infrared sensors for upward obstacle avoidance, ultrasonic sensors, and camera gimbals for stable forward vision. While these drones weigh about $6$-$8$ pounds and cost a few thousand US dollars, they offer the robustness and reliability required for critical applications. Moreover, infrared and thermal cameras~\citep{kumar2011visual} are particularly useful in autonomous human surveillance and rescue operations in darkness and during adverse weather. These sensors provide low-resolution thermal images~\citep{rudol2008human}, which are used to localize moving targets (\textit{e.g.}, people) in darkness. While multiple high-resolution stabilized cameras are useful in these applications, the manufacturers for person-following UAVs tend to avoid using other exteroceptive sensors and try to balance the trade-off between the cost and battery life. 
For instance, although laser scanners are widely used by UAVs for surveying tasks involving mapping and localization~\citep{huh2013integrated,tomic2012toward}, these are not commonly used for person-following applications.

\begin{table*}
\centering
\footnotesize
\caption{Choice of sensors and the corresponding design issues for person-following UAVs.}
\begin{tabular}{|p{1.9cm}||p{1.9cm}|p{2.7cm}|p{4.8cm}|p{3.9cm}|}
\hline
\textbf{Sensor}  & \textbf{Data} & \textbf{Challenges/Limitations}  & \textbf{Usage/Operation} & \textbf{Selected References} \\
\hline \hline
Front Camera  & RGB image & Low visibility; lighting variation  & Computer vision algorithms are used for detection and tracking  & \cite{Skydio,de2015board,pestana2014computer,gaszczak2011real} \\ 
\hline
Vertical Camera & RGB image & Lack of ground textures & Ground textures and visible features are used for stabilization  & \cite{lugo2014framework} \\ 
\hline
RGBD camera &  RGBD data & Presence of sunlight  & In addition to RGB image-space based detection and tracking, distance of the person is approximated from depth-data   & \cite{naseer2013followme,lichtenstern2012prototyping} \\ 
\hline
Ultrasonic sensor & Distance measure & Limited range & Vertical displacement is measured from the distance measures & \cite{lugo2014framework,bartak2015any}  \\ 
\hline
Infrared and Thermal sensor & Thermal image &  Low resolution & Thermal radiation of a person is detected & \cite{kumar2011visual,rudol2008human}  \\ 
\hline
IMU & IMU data & Precision; drift & Flight controllers use the orientation, angular speed, acceleration, and magnetic field information for navigation & \cite{bartak2015any, lugo2014framework}  \\ 
\hline
GPS & Global position, speed, and time & Signal strength; accuracy in indoor settings & The triangulated positioning information is used by the control loop for smart navigation & \cite{Inspire,rudol2008human}  \\ 
\hline
\end{tabular}
\label{AerialSensors}
\end{table*}

Lastly, proprioceptive sensors are used mainly by the flight controller modules. For instance, IMUs measure three-axis rotations and acceleration while the optical-flow sensor measures horizontal (ground) velocity of the UAV. Additionally, ultrasonic and pressure sensors measure altitude and vertical displacements of the UAV~\citep{bartak2015any}. Flight controller modules use these sensory measurements to estimate the UAV pose and eventually control its position and trajectory during flight. Hence, these sensors are critical for the overall successes of the operations. Additionally, advanced UAVs make use of the Global Positioning System (GPS) receivers within the navigation and control loop which allows for smart navigation features such as maintaining a fixed position or altitude. Table~\ref{AerialSensors} summarizes the usage and challenges of different sensors used by person-following UAVs.

\subsubsection{Mode of interaction}
Since the per-episode operating time for UAVs is significantly shorter than that of UGVs and AUVs, their take-offs and landings are frequent. This requires that the person be aware of the UAV's location and available battery at all times in order to facilitate smooth person-following and ease the landing processes. Additionally, for UAVs that are paired to a user application via Wireless Local Area Network (WLAN), the person being followed should not venture outside the WLAN range. Furthermore, the person can positively influence the behavior of the UAV by understanding the underlying algorithms, \textit{e.g.}, by knowing how the UAV navigates around an obstacle, how the rediscovery happens when the target person is lost, etc. While these positive influences via implicit interactions are important for person-following in general, they are more essential in the aerial scenario.

As mentioned earlier, implicit interaction incurs additional cognitive loads on the user. To this end, explicit interactions and commands can simplify the task of controlling the UAV. Most commercial UAVs can be controlled via smart devices~\citep{Skydio,Mavic}, proprietary controllers, wearable beacons, etc. Moreover, hand gesture-based interaction is particularly popular in personal applications where the UAV flies close to the person at a low altitude~\citep{naseer2013followme}. Typical instructions involve changing the robot’s position or initiating a particular task such as start circling around the person, start/stop video recording, make an emergency landing, etc. Nevertheless, hand gesture-based interaction with UAVs can be quite challenging when the UAV flies at a high altitude; these challenges are elaborately discussed later in this paper (in Section~\ref{explicitI}).  

\subsubsection{Granularity}
Similar to the ground and underwater scenarios, a single UAV follows a single person in most commercial and personal applications. Due to the increasing popularity of these applications, research studies have also concentrated largely on this interaction paradigm~\citep{pestana2014computer,lugo2014framework,chakrabarty2016autonomous}. However, a single UAV cannot often fulfill certain application requirements such as capturing an event from different viewpoints or over a long period of time. Hence, critical applications such as search and rescue require multiple UAVs to follow the rescue team~\citep{cacace2016multimodal} and often share a cooperative task (\textit{e.g.}, covering a certain search perimeter). Moreover, a group of cooperative UGVs are more effective for crowd control~\citep{minaeian2016vision} compared to using a single UAV.

While integrating multiple person-following UAVs together can overcome certain limitations of using a single UAV, controlling and interacting with multiple UAVs become increasingly difficult. It increases cognitive load on the users significantly as they need to worry about their battery lives, take-offs and landings, positions, movements, etc. Although it is theoretically possible to interact with multiple UAVs separately and as a group using hand gestures or smart devices, it is not practical for most personal applications. For critical applications, however, UAVs with more advanced autonomous capabilities are used to reduce the cognitive load on the person. In fact, advanced UAVs have features that allow interactions with multiple persons who share the cognitive load in complex operations. For instance, the camera gimbals of Inspire 2.0~\citep{Inspire} can be controlled independently (by a person) while it is interacting with a different person.

\subsubsection{Degree of autonomy}
Unlike ground scenarios, partial autonomy is preferred over full autonomy in most applications for person-following UAVs. The person usually uses a smartphone application for take-offs, positioning, and landing. Then, the UAV switches to autonomous mode and starts following the person. During operation, the person typically uses a smart-phone application or hand gestures to communicate simple instructions for moving the UAV around, taking snapshots, etc. If the UAV loses visual on the person, it hovers until rediscovery is made. UAVs are also capable of emergency landing by themselves if necessary (\textit{e.g.}, when the battery is low or internal malfunctions are detected). These autonomous features minimize cognitive load on the person and reduce the risk of losing or damaging the UAV. 

While partially autonomous UAVs are suitable in controlled settings such as filming sports activities, fully autonomous behavior is suitable in situations where external controls cannot be easily communicated. For instance, autonomous mission planning is essential for applications such as remote surveillance and rescue operations, aiding police in locating and following a fleeing suspect, etc.

\section{State-of-the-art Approaches}\label{sec:approach}
Perception, planning, control, and interaction are the major computational components of an autonomous person-following robotic system. This section discusses these components of the state-of-the-art methodologies and their underlying algorithms. 

\subsection{Perception}\label{sec:approach_Perception}
\begin{figure}[ht]
 \centering
    \includegraphics[width=\linewidth]{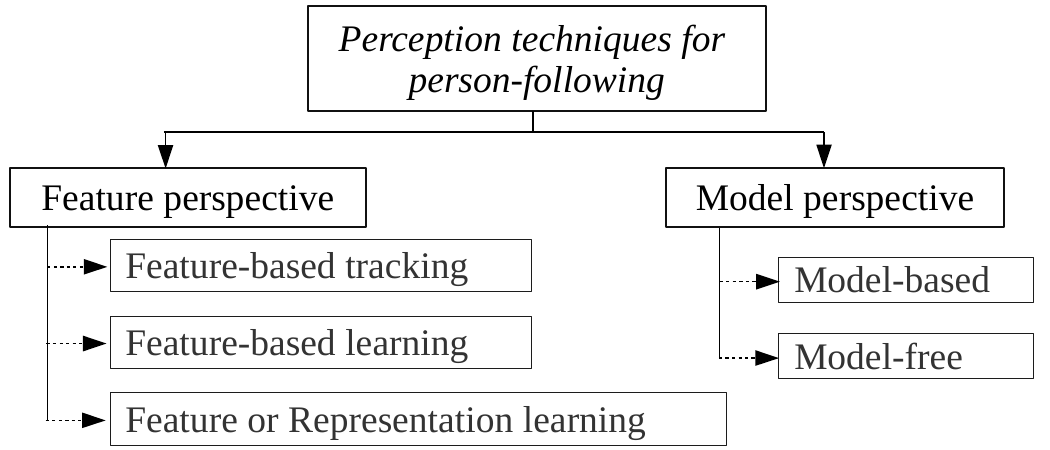}
 \caption{A categorization of various perception techniques for person-following based on the \textit{feature perspective} and \textit{model perspective}.}
 \label{fig:perc}
\end{figure}

An essential task of a person-following robot is to perceive the relative position of the person in its operating environment. The state-of-the-art perception techniques for object following or object tracking, in general, can be categorized based on two perspectives: feature perspective and model perspective (see Figure~\ref{fig:perc}). Based on whether or not any prior knowledge about the appearance or motion of the target is used, they can be categorized into model-based and model-free techniques. On the other hand, based on the algorithmic usage of the input features, perception techniques can be categorized into feature-based tracking, feature-based learning, and feature or representation learning.  

Our discussion is schematized based on the feature perspective since it is more relevant to the person-following algorithms. Additionally, various aspects of using human appearance and motion models are included in our discussion. These aspects, including other operational details of the state-of-the-art perception techniques for ground, underwater, and aerial scenarios are presented in the following Sections.

\subsubsection{Ground Scenario}\label{percep_G}
The UGVs navigate in a two-dimensional (2D) space while following a person (Figure \ref{fig:mixedGround}). Most UGVs adopt a unicycle model~\citep{pucci2013nonlinear} with linear motion along the ground ($XY$) plane and angular motion about vertical ($Z$) axis. One implicit assumption is that the cameras are static and rigidly attached to the robots, as omnidirectional cameras~\citep{kobilarov2006people} are rarely used for person-following applications. The camera feeds and other sensory inputs are fused and sent to the perception module in order to localize the person with respect to the robot. Although the underlying algorithms vary depending on the choice of sensors, they can be generalized into the following categories:

\begin{figure*}[t]
    \centering
    \begin{subfigure}[t]{0.215\textwidth}
        \includegraphics[width=\linewidth]{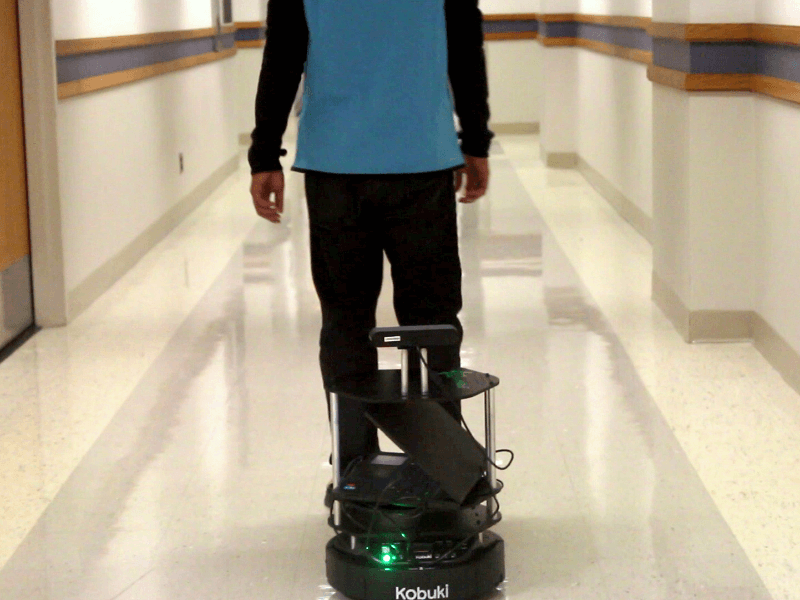}
        \caption{A UGV is following a person from behind}
        \label{fig:mixedGround_a}
    \end{subfigure}%
    ~
    \begin{subfigure}[t]{0.215\textwidth}
        \centering
        \includegraphics[width=\linewidth]{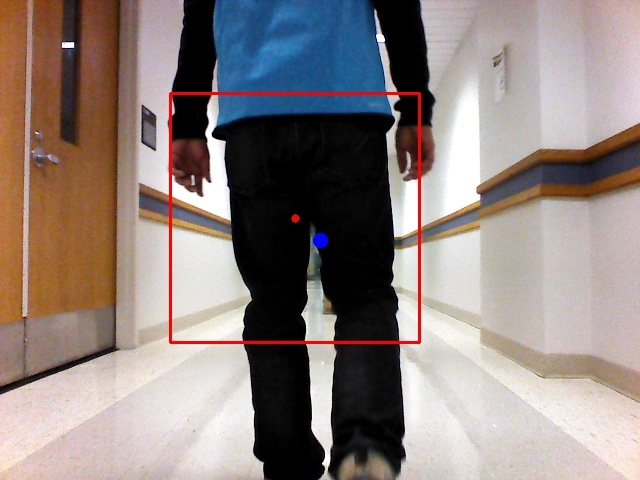}
        \caption{View from the robot's camera (with the detected bounding box) }
        \label{fig:mixedGround_b}
    \end{subfigure}
    ~
    \begin{subfigure}[t]{0.255\textwidth}
        \includegraphics[width=\linewidth]{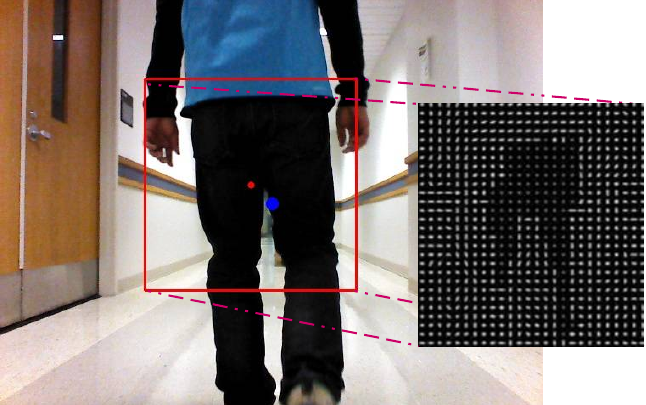}
        \caption{HOG features are shown for the selected region of interest}
        \label{fig:mixedGround_c}
    \end{subfigure}%
    ~
    \begin{subfigure}[t]{0.26\textwidth}
        \centering
        \includegraphics[width=\linewidth]{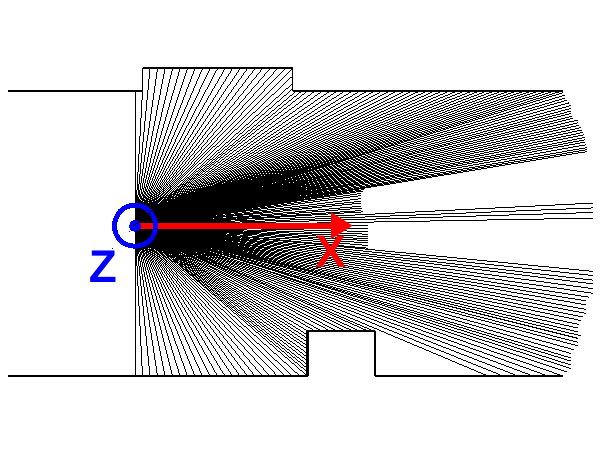}
        \caption{A slice of planar distance measures generated from LRF data (top-view)}
        \label{fig:mixedGround_d}
    \end{subfigure}

    \begin{subfigure}[t]{\textwidth}
        \includegraphics[width=\linewidth]{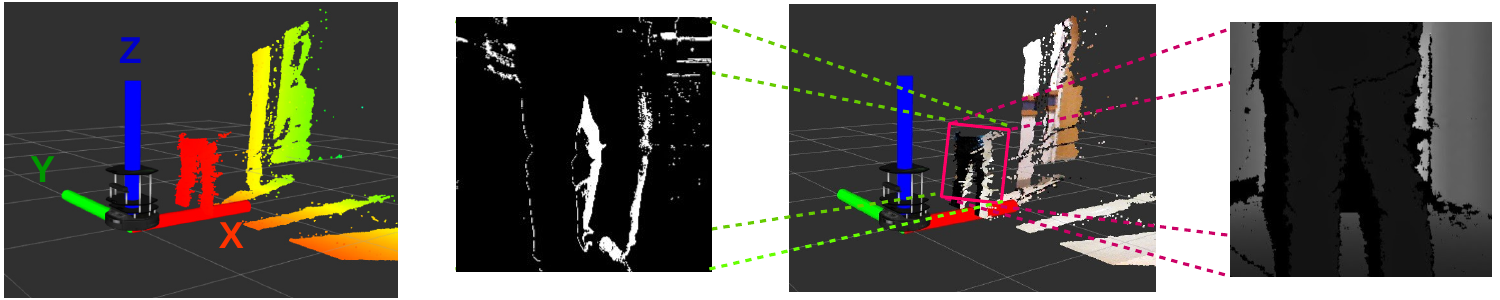}
        \caption{3D point clouds for the global scenario is shown in the left-most image; for a selected region within the robot's field-of-view, the background subtracted binary image and the corresponding depth map is shown on the right}
        \label{fig:mixedGround_e}
    \end{subfigure}%
 
    \vspace{-2mm}
    \caption{Snapshot of a person-following scenario by a UGV, the sensory data for different choices of sensors, and visualizations of the processed data used by various algorithms.}
    \vspace{1mm}
    \label{fig:mixedGround}
\end{figure*}

\subsubsection*{(i) Feature-based Tracking} 
The simplest class of person-following algorithms detect person-specific features in the input feature space. For example, blob detection algorithms use color-based segmentation to track a person in the RGB image space~\citep{hu2009reliable,hu2007robust,schlegel1998vision}. The obvious dependency on specific colors (\textit{e.g.}, person's clothing), make these algorithms impractical for general applications. More robust and portable algorithms can be designed using the depth data generated from an RGBD camera or a stereo camera. As illustrated in Figure \ref{fig:mixedGround_e}, the presence of a person corresponds to a specific pattern in terms of shape, average distance, and the number of points in the 3D point-cloud. Usually, a template is designed based on the expected values of these attributes, which is then used for detection \citep{isobe2014human, satake2013visual}. A family of person-following algorithms applies similar methodologies to laser range finder (LRF) and sonar data. As seen in Figure \ref{fig:mixedGround_d}, slices of planar distance measures from an LRF or directional distance measures from a sonar can be used to detect specific feature patterns or templates pertaining to a person in an upright posture.

More advanced algorithms iteratively refine the initial detection of person-specific features. Mean-shift and particle filter-based algorithms~\citep{germa2009vision,kwolek2004person} are the most popular ones that are used for person-following. Mean-shift performs back-projection to find the probabilities of the target feature map in each point in the feature space. Then, by iteratively following the center of mass of the probability distribution (termed as mean-shift vector), it finds the mode of the distribution that corresponds to the best match for the target feature map. These approaches work very well for unimodal cases and therefore are not very effective in tracking multiple targets. Particle filters, on the other hand, adopt an iterative prediction-update process to derive a set of particles (\textit{i.e.}, candidate solutions). The particles are initialized randomly over the feature space, then iteratively updated based on their similarities with the target feature map. The update-rules and the similarity functions are designed in a way that the particles move toward more prospective regions in the feature space and eventually converge to the target region.

Since searching over the entire feature space can be computationally expensive, it is very helpful to use prior knowledge or to make educated assumptions in order to reduce the search space. For example, Gaussian mixture model-based background subtraction~\citep{stauffer1999adaptive} can be used to avoid searching for person-specific features over the background regions (Figure~\ref{fig:mixedGround_e}). Additionally, exhaustive searching in every frame can be avoided by keeping track of the detected features over sequences of frames. Optical flow-based methods~\citep{handa2008person} and other feature-based trackers~\citep{satake2013visual,satake2012sift} take advantage of this for efficient tracking. Furthermore, educated assumptions on the walking model of a person can also facilitate removing unpromising regions from the feature space~\citep{guevara2016vision}. 

\subsubsection*{(ii) Feature-based Learning} \label{ml_proc}
Another class of approaches makes statistical assumptions on the true underlying function that relates the input feature space to the exact location of the person and then uses machine learning techniques to approximate that function. For example, Histogram of Oriented Gradients (HOG) features are used to train Support Vector Machines (SVMs) for robust person detection~\citep{satake2009robust}. HOG features are histograms of local gradients over uniformly spaced rectangular blocks in the image space. The localized gradient orientations are binned to generate dense feature descriptors. These descriptors, along with other sensory inputs (\textit{e.g.}, depth information) are used to formulate the feature space, which is then used for off-line training of detectors such as SVMs. These detectors are known to be robust and their inference is fast enough for real-time applications. Other supervised models such as decision tree and logistic regression can also be applied by following a similar methodology~\citep{germa2009vision}. Figure~\ref{fig:mixedGround_c} shows HOG features for a particular image patch; as seen, the presence of a person results in a particular spatial pattern in the HOG feature space.

On the other hand, learning models based on Adaptive Boosting (AdaBoost)~\citep{chen2017person} are different in a way that instead of learning a single hypothesis, they iteratively refine a set of \textit{weak} hypotheses to approximate the \textit{strong (optimal)} hypothesis. Use of multiple learners almost always provides better performance than a single model in practice, particularly when the input features are generated using heterogeneous transformations (\textit{e.g.}, linear, non-linear) of a single set of inputs or simply contain data from different sources (\textit{i.e.}, sensors). \cite{dollar2009integral} exploited this idea to extract \textit{integral channel features} using various transformations of the input image. Features such as local sums, histograms, Haar features~\citep{papageorgiou1998general} and their various generalizations are efficiently computed using integral images and then used as inputs to decision trees that are then trained via AdaBoost. A family of these models~\citep{dollar2010fastest, zhu2006fast} is known to work particularly well as pedestrian detectors for near real-time applications such as person-following.

Furthermore, Bayesian estimation and other probabilistic models~\citep{guevara2016vision,alvarez2012feature} are widely used to design efficient person detectors. These models make statistical assumptions on underlying probability distributions of the feature space and use optimization techniques to find the optimal hypothesis that maximizes the likelihood or the posterior probability. One major advantage of these models is that they are computationally fast and hence suitable for on-board implementations.

\begin{figure*}
\centering
\includegraphics [width=\linewidth]{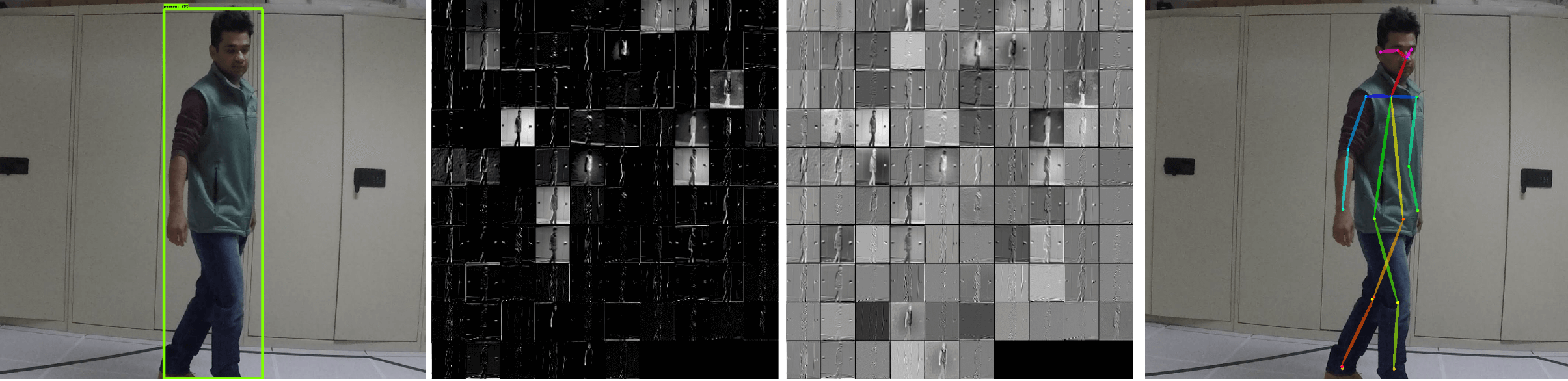}
\vspace{-5mm}
\caption{The left-most image shows a person is detected by a deep object detection model, named Single-Shot Multibox Detector~\citep{liu2016ssd}; visualizations for different feature representations that are extracted at the first two layers of the model, are shown on the next two images; the right-most image shows human body-pose being detected using Open-Pose~\citep{cao2017realtime}. }
\label{fig:conv_ugv}
\end{figure*}

\subsubsection*{(iii) Feature or Representation Learning}\label{deep_proc}
The feature-based learning methods learn optimal hypothesis on a feature space that is designed beforehand from the input sensory data. Consequently, the performance of the underlying algorithm largely depends on how discriminative and informative the feature space is. Deep learning-based approaches try to learn an optimal feature space and the optimal hypothesis simultaneously, which provides a significant boost in performance. Recent developments in Convolutional Neural Networks (CNNs) have made it possible to use these models in real-time applications such as person-following. 

The supervised deep models typically consist of a convolutional network and an additional classification or regression network. The former consists of several convolutional layers that extract the informative features from the input data to generate different feature representations. These feature representations are then fed to a classification or regression network (a set of fully connected layers) for detection. Often a separate region proposal network is used for allowing efficient detection of multiple objects in the scene. Back-propagation and gradient-based optimization techniques are used to find the optimal feature space and optimal hypothesis simultaneously. A sample DeConvnet visualization \citep{zeiler2014visualizing} is provided in Figure~\ref{fig:conv_ugv}, which shows feature representations extracted by a CNN at different layers. Each sub-plot represents the feature-maps that excite a particular neuron in the given layer. Feature-maps for the first and second layer (of a CNN) are shown since they are easier to visually inspect. 
These feature-maps are used by the classifiers and regressors to detect a person and other objects in the scene. 

The CNN-based deep models define the current state-of-the-art for object detection, classification, and visual perception in general~\citep{tfzoo}. However, they require a set of comprehensive training samples in order to achieve good generalization performances by avoiding over-fitting. 
Nevertheless, they often perform poorly in situations such as occlusions, appearance changes of the target (person), random changes in the environments, etc. Online learning schemes~\citep{chen2017integrating} can cope with these issues by adjusting their model-weights based on new observations on the fly. On the other hand, in deep reinforcement learning~\citep{chen2017decentralized} and agent-based models~\citep{gascuena2011agent}, a reward function is defined based on the robot's perceived \textit{state} and performed \textit{actions}. The robot learns sequential decision making to accumulate more rewards while in operation. The overall problem is typically formulated as a Markov decision process (MDP) and the optimal action-state rules are learned using dynamic programming techniques. These methods are attractive because they do not require supervision and they imitate the natural human learning experience. However, they require complex and lengthy learning processes.

Unsurprisingly, the modern person-following robots use deep learning-based person detectors~\citep{jiang2018classification,chen2017integrating,wang2018person} since they are highly accurate in practice and robust to noise, illumination changes, and other visual distortions. More advanced robots go beyond person detection and adopt robust models for human pose estimation and action recognition. These are potentially useful for enabling additional capabilities such as learning long-term human behavior, understanding sentiment, engaging in natural conversation and more, which are attractive for interactive person-following applications in social settings. 

\subsubsection{Underwater Scenario}
As discussed in Section~\ref{sec:UW_scen}, perception is more challenging for underwater diver-following robots. Challenging operating conditions call for two major characteristics of a perception algorithm: robustness to noisy sensory data and fast running-time with limited on-board resources. Consequently, state-of-the-art approaches focus more on robustness and fast running-time than on the accuracy of perception. 

\begin{figure*}[!htb]
    \centering
    \begin{subfigure}[t]{0.48\textwidth}
        \includegraphics[width=\linewidth]{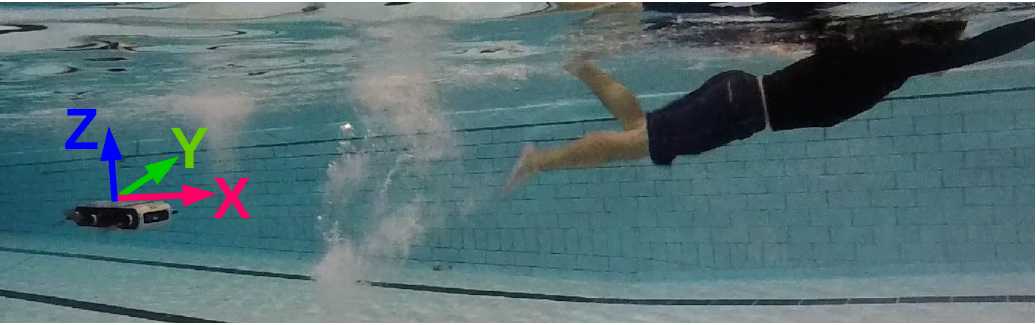}
        \caption{A typical diver-following scenario is shown; the robot moves in a 3D space while following the diver}
        \label{fig:mixedWater_a}
    \end{subfigure}~
    \begin{subfigure}[t]{0.5\textwidth}
        \includegraphics[width=\linewidth]{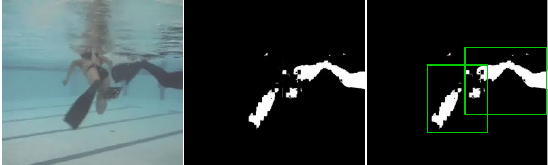}
        \caption{Color-based tracking algorithms perform binary thresholding on image, which is then refined for tracking divers}
        \label{fig:mixedWater_b}
    \end{subfigure}%
    
    \vspace{1mm}
    \begin{subfigure}[t]{0.725\textwidth}
        \centering
        \includegraphics[width=\linewidth]{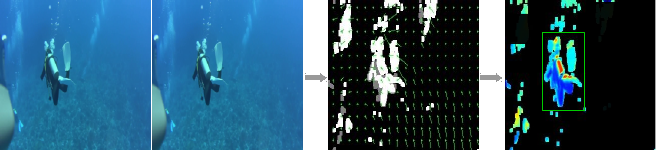}
        \caption{Optical flow-based methods can be used to track a diver's motion; the image in the middle shows the estimated optical flows pertaining to the temporally sequential frames on the left, which is then refined to localize the diver (a heat-map is shown on the right-most image)}
        \label{fig:mixedWater_c}
    \end{subfigure}~
    \begin{subfigure}[t]{0.26\textwidth}
        \centering
        \includegraphics[width=0.95\linewidth]{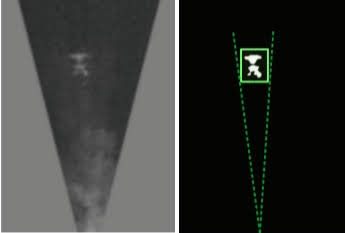}
        \caption{Sonar imagery~\citep{mandic2016underwater} can be refined for tracking divers within a limited range}
        \label{fig:mixedWater_d}
    \end{subfigure}
    
    \vspace{1mm}
    \begin{subfigure}[t]{\textwidth}
        \centering
        \includegraphics[width=0.46\linewidth]{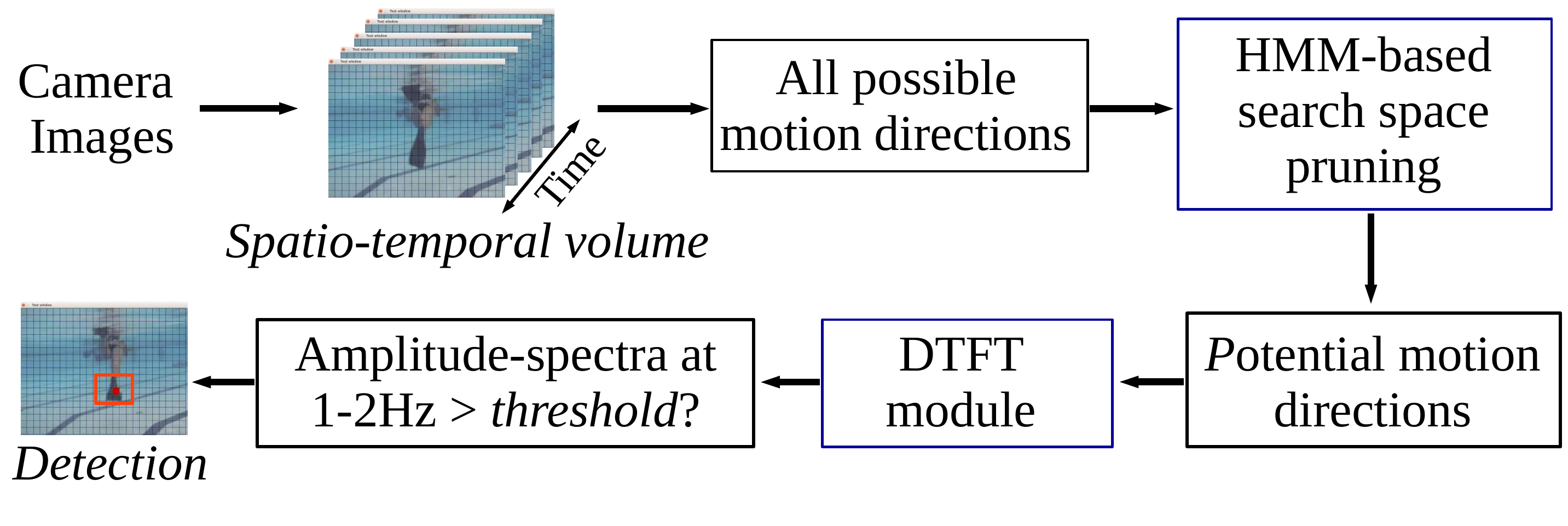} \hspace{3mm}
        \includegraphics[width=0.48\linewidth]{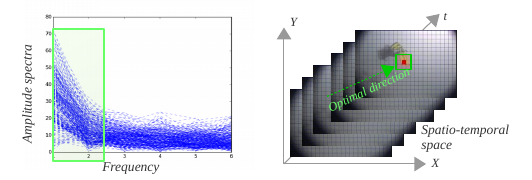}
        \caption{MDPM tracker~\citep{islam2017mixed} uses a Hidden Markov Model (HMM) to track the potential motion directions of a diver, which are then validated using frequency responses of the corresponding motion; as the right plot shows, intensity variations in the spatio-temporal domain along diver's swimming directions correspond to high energy responses on $1$-$2$Hz in the frequency-domain}
        \label{fig:mixedWater_e}
    \end{subfigure}
    
    \vspace{1mm}
    \begin{subfigure}[t]{\textwidth}
        \centering
        \includegraphics[width=\linewidth]{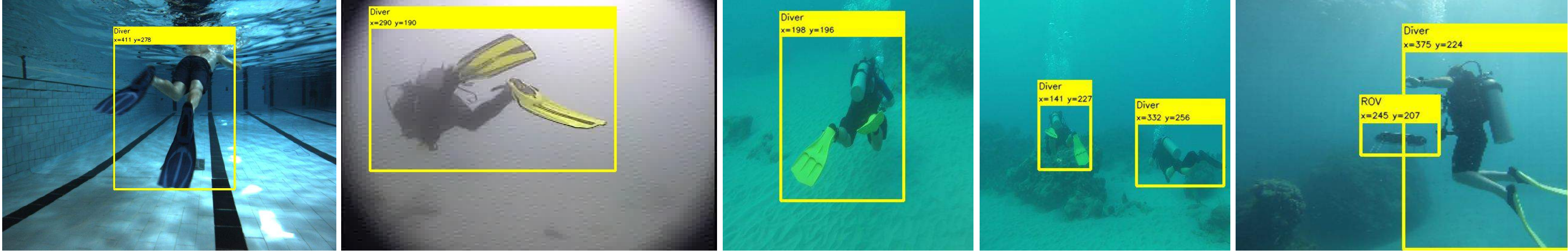}
        \caption{A CNN-based deep model is used to detect divers and other objects (\textit{e.g.}, other robots) by an AUV~\citep{islam2018towards}}
        \label{fig:mixedWater_f}
    \end{subfigure}

    %\vspace{-1mm}
    \caption{Snapshots of autonomous diver-following scenarios and visualization of the processed data used by various algorithms.}
    \label{fig:mixedWater}
\end{figure*}

To this end, simple feature-based trackers are often practical choices~\citep{sattar2006performance}. As illustrated in Figure~\ref{fig:mixedWater}(b), color-based tracking algorithms can be utilized to localize a diver in the image space. These algorithms perform binary image thresholding based on the color of the diver's flippers or suit. The binary image is then refined to track the centroid of the target (diver) using algorithms such as mean-shift, particle filters, etc. AdaBoost is another popular method for diver tracking~\citep{sattar2009robust}; as discussed in Section \ref{ml_proc}, Adaboost learns a strong tracker from a large number of simple feature-based trackers. Such ensemble methods are proven to be computationally inexpensive yet highly accurate in practice. Optical flow-based methods can also be utilized to track a diver's motion from one image-frame to another, as illustrated in Figure~\ref{fig:mixedWater_c}. Optical flow is typically measured between two temporally ordered frames using the Horn and Schunk formulation~\citep{inoue1992robot}, which is driven by the brightness and smoothness assumptions on the image derivatives. Therefore, as long as the target motion is spatially and temporally smooth, optical flow vectors can be reliably used for detection. Several other feature-based tracking algorithms and machine learning techniques have been investigated for diver tracking and underwater object tracking in general. However, these methods are applicable mostly in favorable visual conditions having clear visibility and steady lighting.

Color distortions and low visibility issues are common in deep water scenarios. \cite{sattar2009underwater} showed that human swimming cues in the frequency domain are more stable and regular in noisy conditions than traditionally used spatial features like shape and color. Specifically, intensity variations in the spatio-temporal domain along the diver's swimming direction have identifiable signatures in the frequency domain. These intensity variations caused by the diver's swimming gait tend to generate high energy responses in the $1$-$2$Hz frequency range. This inherent periodicity can be used for robust detection of divers in noisy conditions. Mixed Domain Periodic Motion (MDPM) tracker generalizes this idea in order to track arbitrary motions~\citep{islam2017mixed}. In MDPM tracker, spatial features are used to keep track of the potential motion directions of the diver using a Hidden Markov Model (HMM). Frequency responses along those directions are then inspected to find the most probable one; the overall process is outlined in Figure~\ref{fig:mixedWater_e}. These methods are fast and known to be more robust than simple feature-based trackers.

Use of sonars is effective in unfavorable visual conditions. Sonars use acoustic chirps (low-frequency sound waves) along a range of bearings; then, directional distance measures are calculated from the reflected sound waves. 
AUVs and ASVs (Autonomous Surface Vehicles) most commonly use the \textit{active} sonars to track divers~\citep{mandic2016underwater,miskovic2015tracking} in diver-following applications.
Additionally, the processed sonar image measurements (Figure~\ref{fig:mixedWater_d}) can be fused with the USBL measurements for obtaining reliable tracking estimates at a steady rate. Such sensor fusion increases robustness and works even in cases when either the sonar or the USBL measurements are noisy or unavailable. However, active sonars face challenges in coastal waters due to scattering and reverberation. Additionally, their usage cause disturbances to the marine ecosystem and may also be limited by government regulations on sound levels. Thus, the use of passive sonars such as hydrophones is a practical alternative~\citep{hari2015robust,gemba2014partial}. Passive sonars capture sound waves of diver's breaths and movements, which have inherent periodicity. These waves are then analyzed in the frequency domain to detect periodic bursts of low-frequency sound waves pertaining to the diver's breathing or movements. A similar methodology is used by underwater ROVs that use electric-field sensors to detect the presence of divers within a short range~\citep{lennartsson2009electric}.

Deep learning-based object detection models are recently being investigated for underwater applications as well~\citep{islam2018towards,shkurti2012multi}. 
The state-of-the-art pre-trained models are typically trained (off-line) on large underwater datasets and sometimes quantized and/or pruned in order to get faster inference by balancing the trade-offs between robustness and efficiency~\citep{islam2018towards,islam2018understanding}. As illustrated in Figure~\ref{fig:mixedWater_f}, once trained with sufficient data, these models are robust to noise and color distortions; additionally, a single model can be used to detect (and track) multiple objects. Despite the robust performance, they are not as widely used in practice as in terrestrial scenarios due to their slow on-board running times. 
However, with the advent of mobile supercomputers and embedded parallel computing solutions~\citep{Jetson,coral}, efficient on-board implementations of these models are becoming possible.
Nevertheless, although the online learning and reinforcement learning-based approaches are effective for person tracking when appearance and scene changes~\citep{chen2017integrating,chen2017decentralized}, they are yet to be successfully used in practice for diver-following applications.

\begin{figure*}[!htb]
    \centering
         \begin{subfigure}[t]{\textwidth}
        \includegraphics[width=\linewidth]{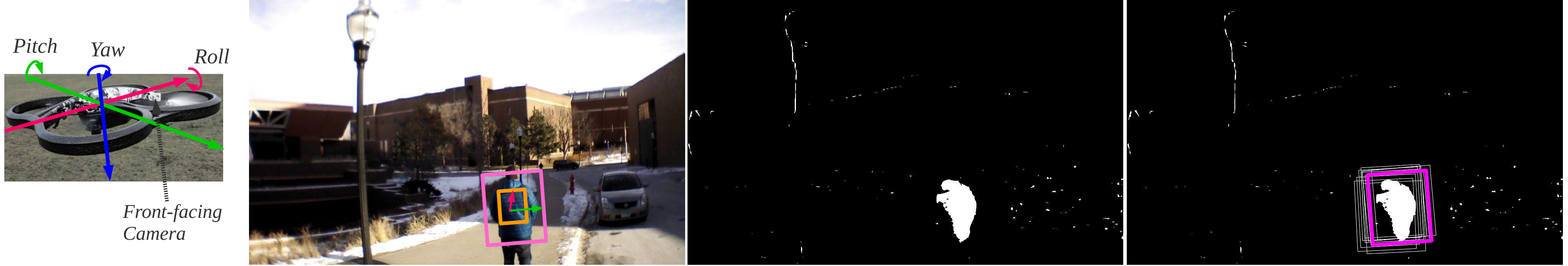}
        \caption{A typical aerial person-following scenario is shown; the UAV moves in a 3D space while following the person. The rightmost figure shows the target bounding-boxes found by the mean-shift algorithm on the color thresholded binary image; the final target bounding-box (marked in pink color) is projected on the original image and refined for tracking}
        \label{fig:mixedAir_a}
    \end{subfigure}
    \vspace{1mm}
 
    \begin{subfigure}[t]{0.49\textwidth}
        \includegraphics[width=\linewidth]{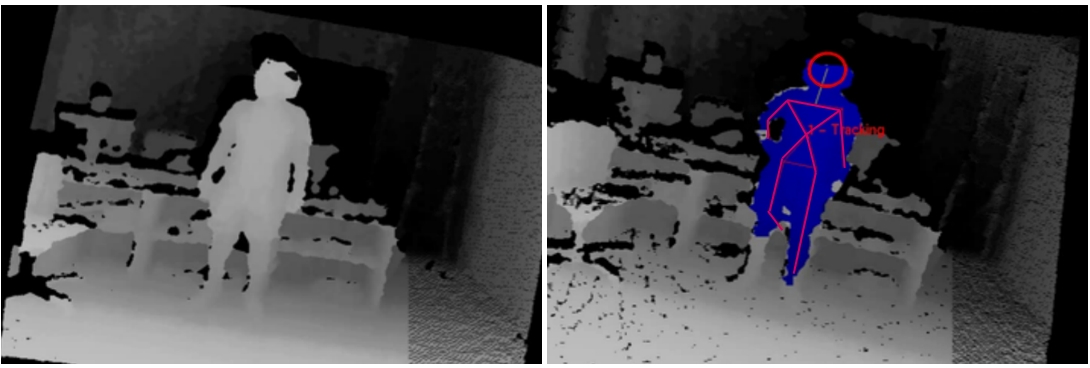}
        \caption{A person being tracked from stabilized depth-map using OpenNI skeleton tracker~\citep{naseer2013followme}}
        \label{fig:mixedAir_b}
    \end{subfigure}~
     \begin{subfigure}[t]{0.5\textwidth}
        \includegraphics[width=\linewidth]{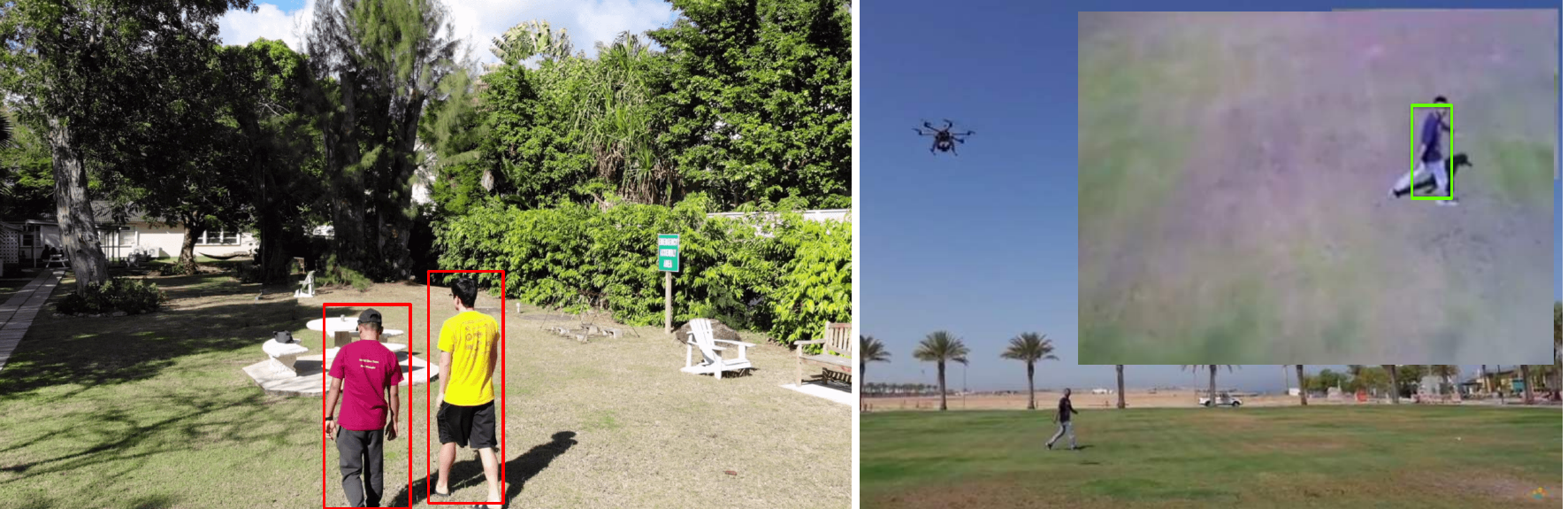}
        \caption{Illustration of different appearances of humans based on the altitude of UAVs; a low altitude case is shown on the left and a higher altitude case is shown on the right~\citep{mueller2016persistent} }
        \label{fig:mixedAir_c}
    \end{subfigure}
    \vspace{1mm}
    
 \begin{subfigure}[t]{0.525\textwidth}
        \centering
        \includegraphics[width=\linewidth]{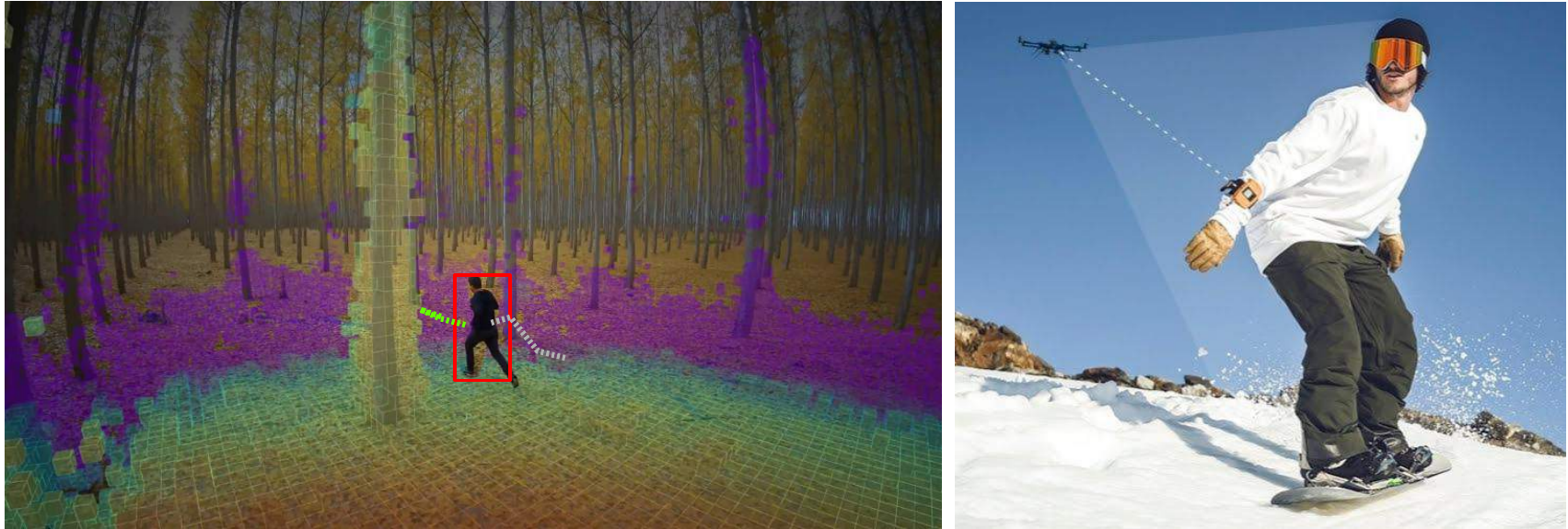}
        \caption{Illustrations of commercial UAVs following an athlete: (left) by creating a 3D map of the surrounding using visual SLAM \& tracking him within the map~\citep{Skydio}; (right) by tracking a paired wrist-mounted device containing a GPS receiver~\citep{Staaker}}
        \label{fig:mixedAir_d}
    \end{subfigure}~
 \begin{subfigure}[t]{0.465\textwidth}
        \includegraphics[width=\linewidth]{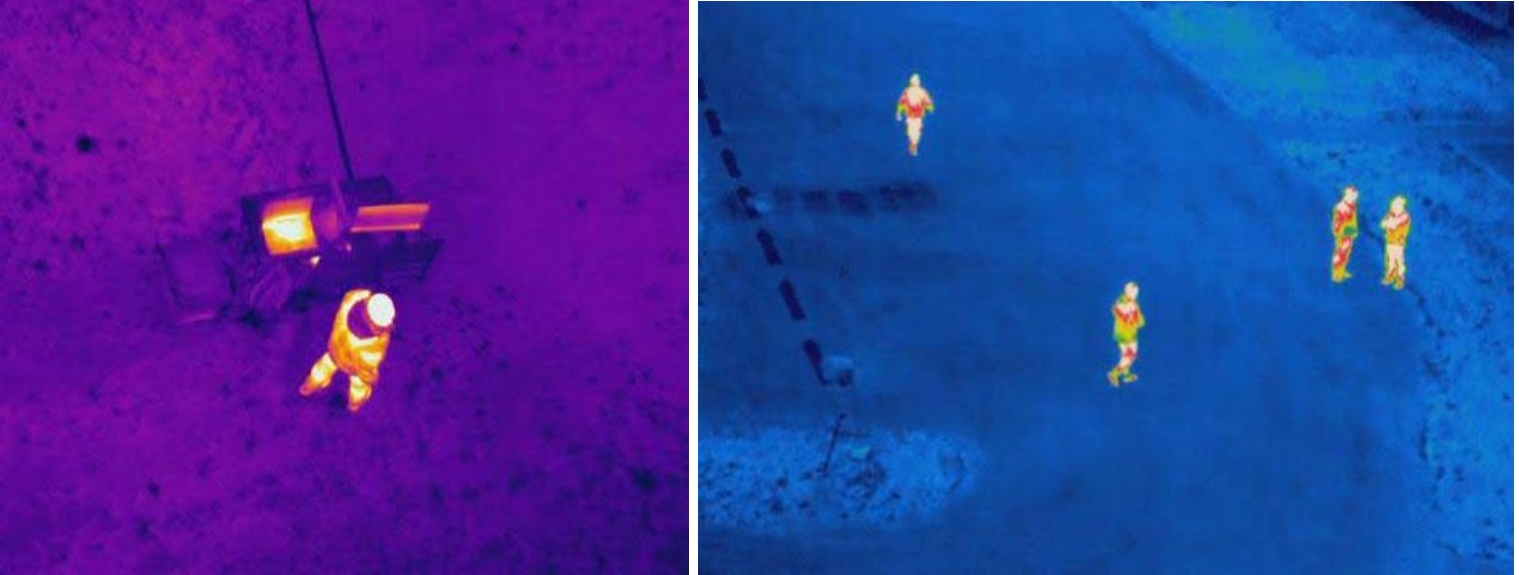}
        \caption{Snapshots taken by a military-grade thermal infrared camera~\citep{spi}; these images capture the infrared energy reflected by an object (such as humans), which can be used to isolate them from the background}
        \label{fig:mixedAir_e}
    \end{subfigure}
    
    \vspace{-1mm}
    \caption{Snapshots of aerial person-following scenarios by UAVs and visualization of the processed sensory data used by various algorithms.}
    \label{fig:mixedAir}
\end{figure*}

\subsubsection{Aerial Scenario}
The underlying mechanism of a perception algorithm for person-following UAVs is mostly defined by two aspects: the expected flying trajectory of the UAV and the available sensory data. For instance, in some personal applications, the UAV flies close to the person at a low altitude (\textit{e.g.}, $4$-$6$ meters from the ground). The perception algorithms in such a setup are similar to those in ground scenarios, as illustrated in Figure \ref{fig:mixedAir_a}. On the other hand, applications such as filming sports activities demand rather complex trajectories of the UAV while following an athlete (Figure \ref{fig:mixedAir_d}). Robust detection and control mechanisms are required in these setups including reliable sensing. Lastly, autonomous surveillance and rescue operations involve sensing from long distances, often in unfavorable sensing conditions; hence, perception techniques differ among these scenarios.

Standard feature-based tracking methods are suitable if the UAV is expected to fly close to the person while maintaining a smooth horizontal trajectory. As seen in Figure \ref{fig:mixedAir_a}-\ref{fig:mixedAir_b}, the camera image captures most of the person's body with no significant perspective distortions; hence, computer vision-based object detectors and pedestrian detectors perform well in such scenarios. To this end, color-based segmentation, mean-shift, particle tracker, and HOG-based detectors are widely used~\citep{kumar2011visual,teuliere2011chasing,lugo2014framework,higuchi2011flying}. The operations of these algorithms are already discussed in Section~\ref{percep_G}. 
In a seminal work,~\cite{pestana2014computer} showed that OpenTLD trackers can achieve robust performance for outdoor suburb environments. TLD trackers decompose a target-tracking task into tracking, learning, and detection; they are known to perform well for long-term object tracking in general. One limitation of TLD trackers is that they often incorporate the background into the learning over time, leading to quick target drift. 

\begin{figure*}
\centering
\includegraphics [width=0.95\linewidth]{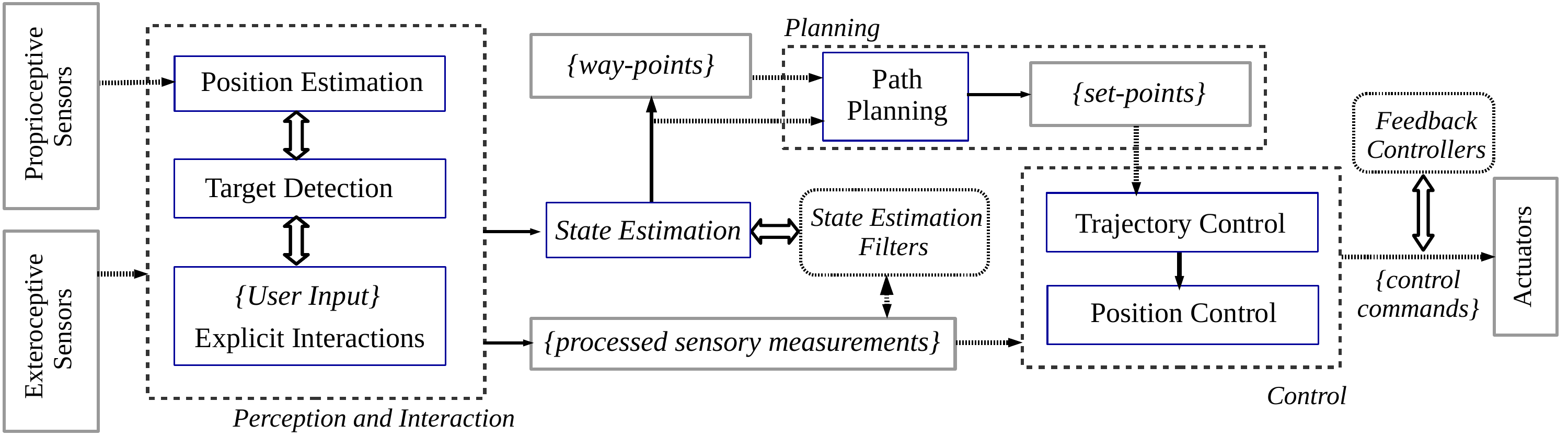}
\vspace{-2mm}
\caption{An outline of the data and control flow among major computational components of a person-following system; this flow of operation is generic to most autonomous object-following paradigms.}
\label{fig:ctr}
\end{figure*}

More reliable sensing and robust tracking performance can be achieved by using an additional depth sensor (\textit{e.g.}, RGBD camera), particularly for indoor applications.~\cite{naseer2013followme} presented an indoor person-following system using two cameras; a regular camera for determining the 3D position of the UAV based on markers on the ceiling and a depth camera to detect a person in 3D. The images from the depth camera are warped based on the calculated 3D position. The stabilized depth images are then used for robust perception using the OpenNI platform (Figure \ref{fig:mixedAir_b}).~\cite{gioioso2014flying} also used an RGBD camera to detect hand gesture-based teleoperation commands for UAVs.  Such systems, however, are limited to indoor environments and small motions. Additionally, they often require a remote computer for intensive computations.

For challenging outdoor applications where the UAV trajectory changes rapidly due to dynamic obstacles or fast movements, the person may appear significantly different from different viewpoints. Hence, the perspective distortions need to be taken into account.~\cite{de2015board} used \textit{ground plane estimation} techniques to approximate the orientation of the ground plane in 3D relative to the position of the camera. Then, object heights in the image are predicted based on the homography of the ground plane and the real-world sizes of the objects. They exploited this idea to localize prospective rectangular regions in the image space for detecting pedestrians of expected height between $160$cm and $185$cm. This allows approximation of the height of the person in different perspectives and thus reduces the search space, which leads to efficient tracking performances. They used Aggregate Channel Feature (ACF)-based standard pedestrian trackers and achieved good on-board performances. A number of other online tracking algorithms are investigated by~\cite{mueller2016persistent} for person-following and general object tracking by UAVs. 
They also presented a \textit{camera handover} technique where one UAV can pass the tracking task over to another UAV without interruption, which can be useful in long-term tracking and filming applications. 
Moreover, some commercial UAVs build a 3D map of the surroundings using techniques such as visual SLAM (Simultaneous Localization and Mapping) and follow their target (person) within the map~\citep{Skydio}. These UAVs are usually equipped with advanced features such as obstacle avoidance, target motion prediction, etc. Furthermore, UAVs that capture sports activities often track the location information provided by a paired controller device which is carried by or mounted on an athlete, as illustrated in Figure~\ref{fig:mixedAir_d}. 
The paired device is equipped with a GPS receiver and it communicates information related to motion and orientation of the athlete; this additional information helps the UAV plan its optimal trajectory for filming~\citep{vasconcelos2016person}.     

As mentioned, the thermal and IR cameras are particularly useful in autonomous surveillance, search and rescue, and other military applications. Thermal images project the emissions of heat from various objects in the image space, which can be easily identifiable from its surroundings. This feature is vital while sensing from long distances and in unfavorable sensing conditions. Figure~\ref{fig:mixedAir_e} shows snapshots of a military-grade thermal infrared camera~\citep{spi}; as seen, the warm objects can be easily located in the image. In fact, one major advantage of using thermal imagery is that simple computer vision techniques can be used for robust detection. For instance,~\cite{portmann2014people} showed that the standard background subtraction techniques can be used to segment regions that are both hotter and colder than the environment. Then, HOG-based detectors or particle filters can be used to track humans in the segmented regions.~\cite{gaszczak2011real} used the mean-shift algorithm on the background-subtracted thermal image and achieved good results. Additionally, they showed that the Haar classifiers can be used to detect human body signatures as well as other objects accurately. These methods are computationally inexpensive and suitable for on-board implementations.

Lastly, deep learning-based person detectors are yet to be explored in depth for the aerial applications, largely due to the limited on-board computational resources available, particularly in consumer-grade UAVs. Nevertheless, some of the recent commercial UAVs such as Skydio R1~\citep{Skydio} use mobile embedded supercomputers in order to use deep visual models in real-time. It is safe to say that with faster mobile supercomputers and better low-power computing solutions, efficient on-board implementations of various deep learning-based perception modules will soon become possible and they will be more commonly used by person-following UAVs in the near future.

\begin{figure*}
\centering
\includegraphics [width=0.95\linewidth]{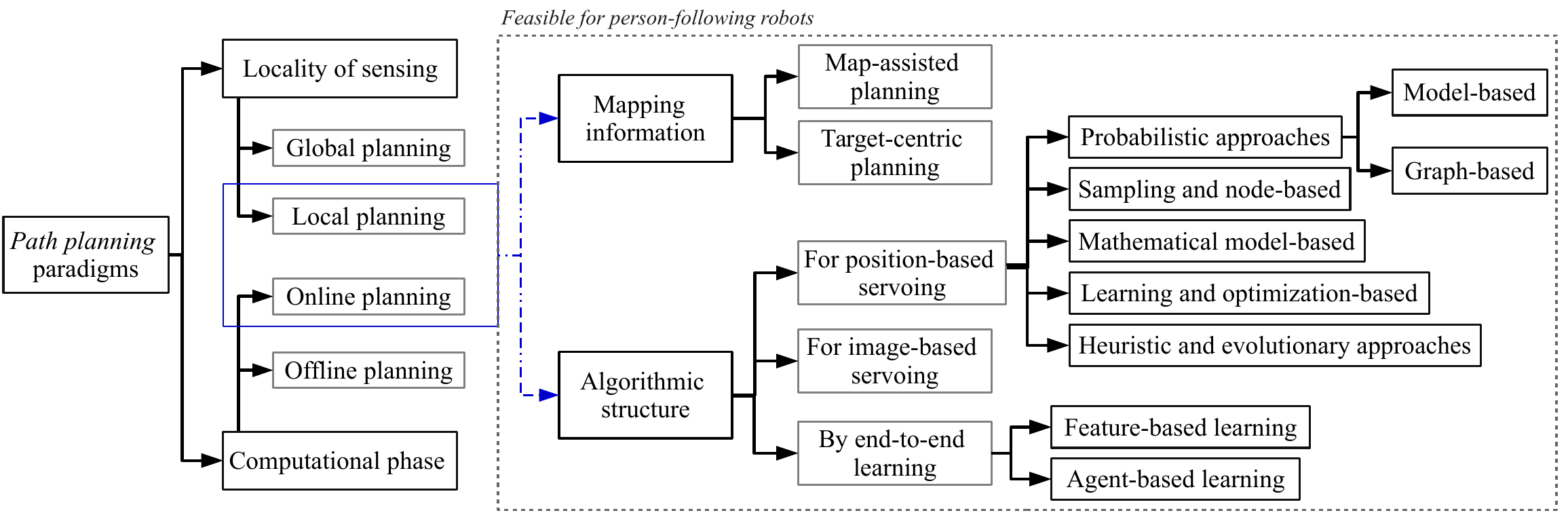}
\vspace{-1mm}
\caption{{  Categorization of path planning algorithms from the perspective of sensing, methodology, and computation.}}
\label{fig:plan1}
\end{figure*}

\subsection{Planning and Control}\label{sec:plan_ctr}
Once the perception module gets an estimate of the target (person) pose by processing the sensory inputs, control commands need to be generated in order to achieve the desired motion. The ground robots navigate in a 2D plane whereas the underwater and aerial robots navigate in 3D spaces; hence, the corresponding control signals and their operational constraints vary. However, the overall operation is mostly similar; Figure~\ref{fig:ctr} illustrates an outline of the operational flow for an autonomous person-following system. The following discussions provide an overview of the planning and control modules that are standard for general object following, and focuses on the aspects that are particularly important for person-following applications.

First, the target person's position and heading information are estimated with respect to the robot's known frame of reference. Additionally, the sensory data is processed and sent to the state estimation filters. These \textit{observed measurements} are used by the filters to refine the state estimation through iterative \textit{prediction-update} rules. Linear quadratic estimation such as Kalman Filter (KF)~\citep{kalman1960new} and non-linear estimation such as Extended Kalman Filter (EKF)~\citep{julier1997new} are most widely used for this purpose. Unscented Kalman Filter (UKF)~\citep{wan2000unscented} addresses the approximation issues of the EKF and is often used in practice for state estimation from noisy sensory data. Methodological details of these algorithms are beyond the scope of this paper; interested readers are referred to \cite{jung2012control,yoon2014real,lugo2014framework,morioka2012control,satake2009robust,sattar2008enabling,teuliere2011chasing}.

The refined measurements are then processed to generate a set of \textit{way-points}, \textit{i.e.}, a representation of potential trajectories for the robot in order to follow the target person. The path planner uses this information and finds the optimal trajectory by taking into account the estimated relative positions of the static obstacles, other humans, and dynamic objects in the environment.  The constraint here is to optimize some aspect of the anticipated motion of the robot such as travel-time, safety, smoothness of the motion, etc. A sequence of points are then generated to discretize the anticipated motion; this set of points pertaining to the optimal trajectory is generally termed as the \textit{set-points}~\citep{doisy2012adaptive}. Finally, the control modules analyze the set-points and generate navigation commands to drive the robot. The generated navigation commands are usually fed to a set of feedback (\textit{e.g.}, PID) controllers. This process of robot control is also generic for most applications and interested readers are referred to \cite{de2012theory,pounds2010modelling,mezouar2002path} for further details.

Figure~\ref{fig:plan1} illustrates a categorical overview of various types of path planning algorithms in the literature. Based on the locality of sensing, planning can be either \textit{global} (for fully observable environments)  or \textit{local} (for partially observable environments). Additionally, if the optimal path (to target) is computed first and then executed sequentially, it is termed as \textit{offline} planning; on the other hand, the planned path is refined dynamically in \textit{online} planning paradigms. Since person-following robots are deployed in partially observable and dynamic environments, they require local and online path planning in order to adapt to the irregular and unpredictable changes in their surroundings. 
Path planning algorithms for person-following robots can be further categorized based on mapping information and on their algorithmic structure. 
Although these algorithms are fairly standard for dynamic target-following, a brief discussion on their operational considerations for person-following is presented in the following Sections.

\subsubsection{Map-assisted vs Target-centric Planning}
\begin{figure}[h]
\centering
\includegraphics [width=\linewidth]{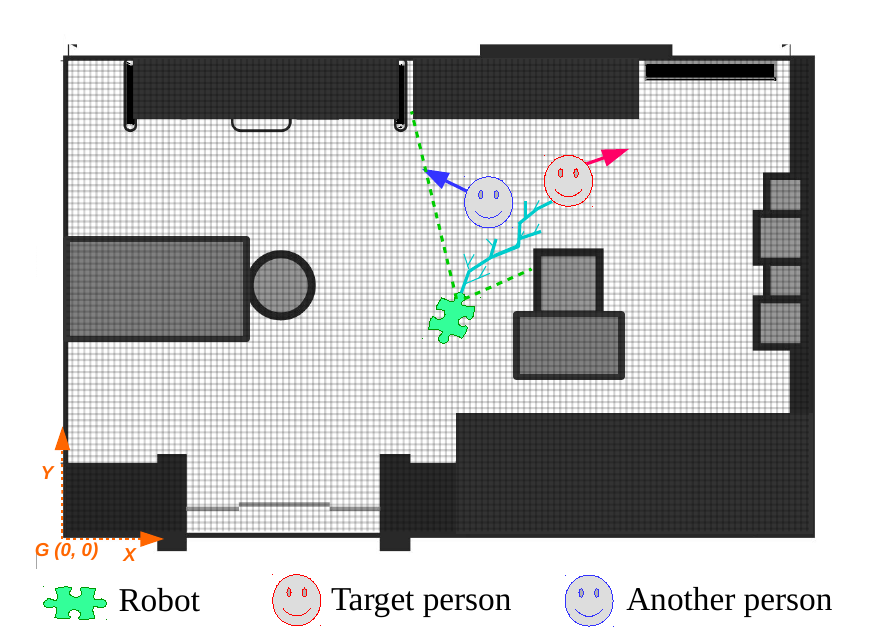} 
\vspace{-5mm}
\caption{A simplified illustration of map-assisted 2D path planning of an ground robot by avoiding static and dynamic obstacles within the map.}
\label{fig:plan3}
\end{figure}

Map-assisted planning is feasible for structured environments with a known map, particularly for person-following ground robots in indoor settings~\citep{nikdel2018the,doisy2012adaptive}. The global map of the environment (including static objects) is used as prior knowledge. {Typically a \textit{static planner} keeps track of the robot's location within the map and its admissible waypoints by taking into account the static obstacles in the environment~\citep{ahn2018formation}. The \textit{dynamic planner} then refines these waypoints by considering the motions of the dynamic objects in the environment (Figure~\ref{fig:plan3}).} 
Other constraints such as social awareness, implicit interactions, etc., can also be considered in the refinement process that eventually generates the optimal path~\citep{cosgun2016anticipatory}. Then, standard map-based navigation techniques are used to invoke the person-following motion.

\begin{table*}[ht]
\centering
\footnotesize
\caption{Various classes of path planning algorithms used by person-following robots for position-based servoing (based on the categorization shown in Figure \ref{fig:plan1}), and their operational consideration. }
\begin{tabular}{|p{1.6cm}||p{5.1cm}|p{3.2cm}|p{2.8cm}|p{2.4cm}|}
\hline
\textbf{Categories}  & \textbf{Operation} & \textbf{Constraints}  & \textbf{Advantages} & \textbf{Selected References} \\ \hline \hline
%Global planning & The total source-to-goal planning is done before the robot starts its motion & The environment (ideally static) has to be completely observable/known  & Good at finding the optimal path with reliable uncertainty bounds & \cite{tsardoulias2016review,zhu2015global,hwang1992potential} \\   \hline 

%Local planning & Planning is done dynamically while the robot is moving & Reliable local sensing is required at a good rate & Capable of responding to dynamic environmental changes & \cite{montiel2015path,hebecker2015model,stentz1994optimal,sedighi2004autonomous} \\ \hline \hline
Probabilistic approaches & The planning hypothesis given the constraints is represented as a generative or inference model & The probabilistic assumptions on the system model might not always hold & Computationally fast and good for planning with limited sensing & \cite{sung2016hierarchical,park2013autonomous} \\ \hline 

Sampling and node-based &  The workspace is sampled into nodes, cells, grids, or potential fields; then, different search methods are used to find the optimal path & Prior knowledge about the environment is needed & Good at finding sub-optimal and often optimal solutions  & \cite{huskic2017outdoor,triebel2016spencer,hoeller2007accompanying} \\ \hline 

Mathematical model-based & The environment is modeled as a time-variant kino-dynamic system; then minimum cost path or maximum utility path is computed & Can be computationally expensive; analytic solution of the system may not exist & Reliable and optimal & \cite{cosgun2013autonomous,doisy2012adaptive,tisdale2009autonomous} \\ \hline 

Learning and optimization-based & Parameters of the optimal planning hypothesis are approximated using machine learning models  & Requires accurate feature representation and rigorous training process  & Can easily deal with complex and dynamic environments & \cite{triebel2016spencer,morales2004machine} \\ \hline 

Heuristic and evolutionary approaches & Planning hypothesis is evaluated using a heuristic or bio-inspired objective function; then, iteratively searched for optimal solution   & Search space can be very large; often produce locally optimal solution & Good at dealing with complex and dynamic environments & \cite{gong2011multi,sedighi2004autonomous} \\ \hline

%Online planning & Planning is refined dynamically until the goal is achieved & Reliable sensing is required at a good rate & Can react and correct planning behavior in dynamic environments & \cite{zadeh2016auv,benavidez2011mobile,stentz1994optimal} \\ \hline 

%Offline planning & Planning is performed once per epoch, the optimal path is saved in memory and executed sequentially   & Cannot react to dynamic changes in the environment; prone to target drift & Low reactive operational overhead; ideal for static and known environments & \cite{marin2018global,zhu2015global,raja2012optimal}  \\ 
\hline
\end{tabular}
\label{PlanningAlgo}
\end{table*}

\begin{figure}
\centering
\includegraphics [width=\linewidth]{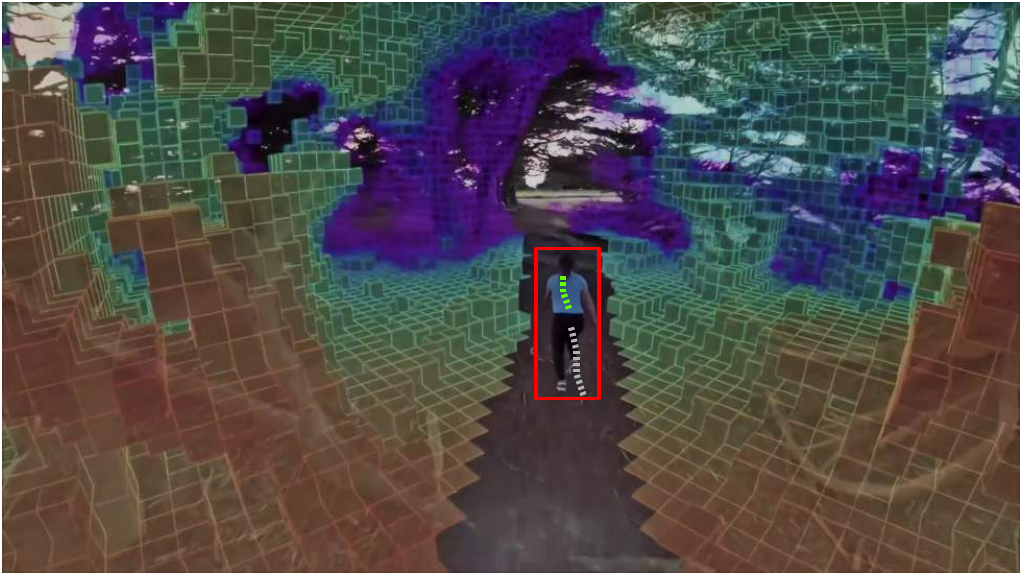} 
\vspace{-5mm}
\caption{A UAV is using a multi-camera depth map of the partially observed environment for target- centric planning in order to follow the person~\citep{Skydio}.}
\label{fig:plan4}
\end{figure}

Although a global map can significantly simplify the planning and navigation processes, it is usually not available in outdoor applications. In such cases, a target-centric approach is adopted. First, the locally sensed information is used to create a partial map of the environment; traditional SLAM-based techniques are most commonly used for this purpose~\citep{ahn2018formation,cosgun2016anticipatory}. As illustrated in Figure~\ref{fig:plan4}, the UAV creates a 3D (depth) map of the partially observed environment in order to find the optimal path for person-following~\citep{Skydio}. Such \textit{reactive} path planning sometimes leads to non-smooth trajectories, particularly if the person moves fast in a zigzag trajectory~\citep{tarokh2010vision}. Anticipatory planning, \textit{i.e.}, predicting where the person is going to be next and planning accordingly, can significantly alleviate this problem and thus widely used in practical applications~\citep{nikdel2018the,tarokh2014vision,hoeller2007accompanying}.

\subsubsection{Planning for Position-based Servoing}
In position-based servoing, path planner finds the optimal path to follow a target using its estimated position with respect to the robot. For instance, a person-following UAV uses its current 3D position as the \textit{source}, estimated 3D location of the person as the \textit{destination}, and then uses source-to-destination path planning algorithms to find the optimal path that meets all the operational constraints. It is to be noted that this planning can be either map-assisted or target-centric, depending on whether or not the global mapping information is available.

Standard path planners typically represent the state space using cells, grids, or potential fields and then apply various search methods to find the optimal source-to-destination path. For instance, the navigation space (and the locations of relevant objects) is often interpreted by an \textit{occupancy grid}, and graph search-based algorithms such as A$^*$, D$^*$, IDA$^*$ (Iterative Deepening A$^*$), etc. are used to find the optimal path~\citep{ahn2018formation,huskic2017outdoor,muller2008socially}. Another approach is to randomly sample the state space and attempt to establish source-to-destination connectivity using techniques such as Rapidly-Exploring Random Trees (RRT)~\citep{triebel2016spencer}, RRT$^*$, Probabilistic Road-Maps (PRM)~\citep{hoeller2007accompanying}, etc. These methods are good at finding near-optimal solutions at a fast rate in large search spaces where ensuring completeness is computationally expensive; hence, they are widely used in real-time path planning for person-following robots.

It is also common to represent the planning hypothesis given the constraints as a probabilistic inference model. Then, the problem reduces to finding a minimum cost or maximum utility path from the search space of all admissible paths. Machine learning models, heuristic and evolutionary approaches are also used to approximate the optimal solution (\textit{i.e.}, to find a near-optimal path), particularly if the search space is too large~\citep{triebel2016spencer,gong2011multi}. Moreover, the problem can be modeled as a Partially Observable Markov Decision Process (POMDP) in order to perform online planning in a continuous state, and action space~\citep{goldhoorn2014continuous,triebel2016spencer}. POMDPs are good at dealing with dynamic environments and complex agent behaviors. However, they can be computationally intractable and generate sub-optimal solutions. Therefore, approximate solutions are typically formulated with an assumption of a discrete state, and/or action space.

Table~\ref{PlanningAlgo} summarizes the different classes of path planning algorithms for position-based servoing and highlights their operational considerations in person-following applications. 
These algorithms are fairly standard; interested readers are referred to~\cite{gonzalez2015review,yang2016survey} for further methodological details.

\subsubsection{Planning for Image-based Servoing}\label{sec:plan_viz_serv}
\begin{figure}[t]
 \centering
    \includegraphics[width=\linewidth]{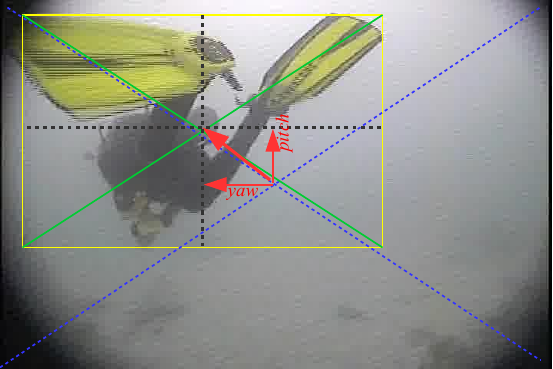}\vspace{-1mm}
 \caption{{Illustration of a bounding-box reactive planner; the horizontal and vertical displacements of the center of the detected bounding box is used for image-based servoing.}}
 \label{fig:serv}
\end{figure}

Autonomous navigation of a robot using visual feedback is known as image-based (visual) servoing, where the path planner uses image-based features in order to find the optimal path to follow the target~\citep{gupta2017novel}. Image-based servoing is particularly useful when it is difficult to accurately localize the target, particularly underwater and in GPS-denied environments~\citep{pestana2014computer}. For instance, AUVs often use \textit{bouding-box reactive} path planners for diver-following~\citep{islam2018towards}. Here, the planning objective is to keep the target diver at the center of the robot's view. That is, motion commands are generated in order to bring the observed bounding box of the diver to the center of the camera image (Figure~\ref{fig:serv}). The relative distance of the diver is approximated by the size of the bounding box and forward velocity rates are generated accordingly. In addition, the yaw and pitch commands are normalized based on the horizontal and vertical displacements of the observed bounding box-center from the image-center; these navigation commands are then regulated by the controller to drive the robot.

\begin{figure}[hb]
 \centering
 \includegraphics[width=\linewidth]{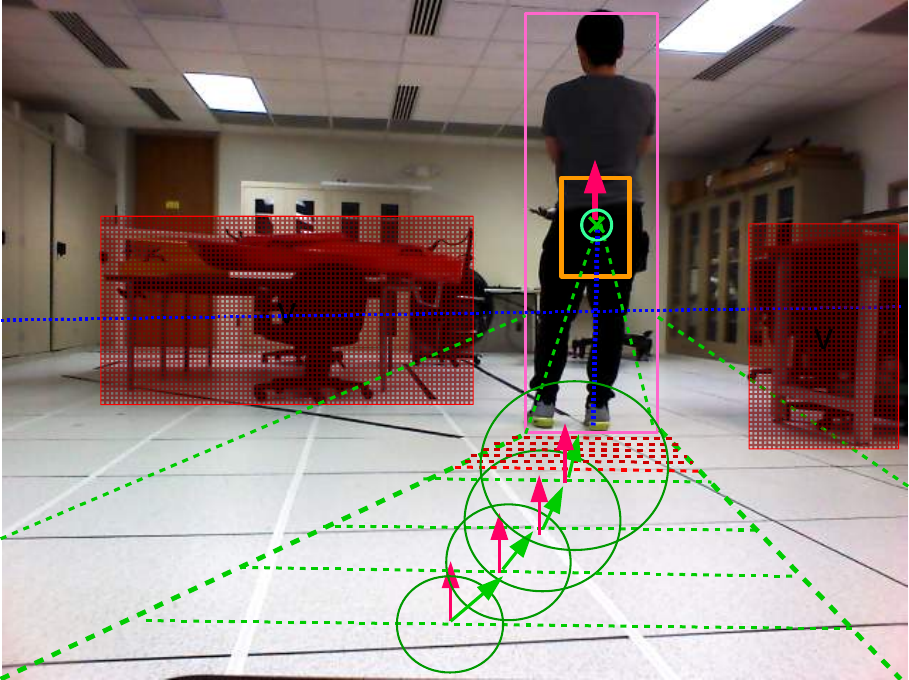}\vspace{-1mm}
 \caption{{  Illustration of a simple planning strategy for a UGV; it finds a straight-line trajectory in order to remain immediately behind the person while avoiding obstacles.}}
 \label{fig:plan2}
\end{figure}

Furthermore, it is common to simplify the planning component for image-based servoing in order to reduce computational burdens on the robot. For instance, diver-following robots sometimes plan a straight-line trajectory to remain immediately behind the diver~\citep{islam2017mixed}. A similar strategy is adopted by ground robots as well~\citep{wang2018accurate,doisy2012adaptive,brookshire2010person}, with an additional planning component for obstacle avoidance. As illustrated in Figure~\ref{fig:plan2}, person-following UGVs can use tools from prospective geometry to get the relative homography of the orthogonal planes and then find the optimal path along the ground plane by keeping safe distances from the person and obstacles. This simplifies the operational complexities and is often sufficient for non-critical applications.

\subsubsection{Planning by End-to-end Learning}
End-to-end techniques try to \textit{learn} problem-specific robot navigation rules directly from input sensory data. 
This way of coupling the perception, planning, and control modules together is inspired by the \textit{self-driving-car} concept and is very popular these days. 
Several deep learning-based models for supervised learning and agent-based reinforcement learning are recently introduced for person-following as well~\citep{dewantara2016generation,goldhoorn2014continuous}. Typically these models are first trained in simulations with an existing motion planner and then transferred to real-world environments for further tuning. Researchers have reported exciting results demonstrating their effectiveness for UGVs in autonomous 2D navigation, avoiding obstacles, following people in near-optimal paths, multi-agent cooperation, etc. However, these techniques are applied mostly for person-following UGVs in indoor settings, and sometimes only in simulation environments~\citep{pierre2018end,dewantara2016generation}. Therefore, more research attention is needed in order to improve and generalize these techniques for a wide range of other person-following applications.

\subsubsection{Other Considerations for Planning and Control}\label{reid_p}
In addition to the operating constraints for person-following mentioned already, there are other practical, and often application-specific considerations for effective planning and control. Several such aspects are discussed in this Section.

\textbf{Planning ahead to avoid occlusion:} 
The most essential feature of a person-following robot's planning module is to ensure that the target person is in the field-of-view during its motion. The trajectory needs to be planned in a way that in addition to meeting the standard criteria of an optimal path, \textit{e.g.}, based on the distances from obstacles, expected travel time, smoothness of anticipated motion, etc., the person remains reasonably close to the center of its field-of-view and not occluded by obstacles. This is challenging if the sensing range is limited, and especially in the presence of dynamic obstacles.
Typically, a probabilistic map~\citep{hoeller2007accompanying} for motions of the moving objects in the scene is formulated and then path planning is performed on the dynamic occupancy field; a temporal window of \textit{motion history} is maintained to facilitate such formulation. Another approach is to predict the positions and velocities of the moving objects a few time-epochs into the future and plan the optimal trajectory ahead of time. Such anticipatory planning is particularly important for person-following UGVs which are meant to stay ahead of the person~\citep{mi2016system}, and UAVs that film fast-moving athletes~\citep{Skydio}.

\textbf{Camera control:} 
If the person-following robot is equipped with a pan-and-tilt camera, a controller module is required to ensure that the camera is always pointed to the person during navigation~\citep{doisy2012adaptive,tarokh2010vision}. In addition, it is common for person-following UAVs to have camera-gimbals; if so, an additional module is required to control the gimbals' angles instantaneously~\citep{Skydio,Staaker} based on the person's relative motion.

\textbf{Person re-identification and recovery}:  
In addition to robust person detection and tracking, person-following robots need to be able to plan to re-identify when necessary~\citep{koide2016identification}. Moreover, these techniques are essential for accompanying a specific person~\citep{eisenbach2015user,ilias2014nurse}. Predictive and probabilistic models such as Kalman filters, particle filters, etc., are typically used to estimate the person's location in future, which can be used as prior knowledge in case of a \textit{missing target} situation. That is, when the robot fails to detect the person (due occlusion or noisy sensing), the \textit{recovery planner} can use that anticipated person's location as prior and search probable locations for re-identification~\citep{do2017reliable,gupta2017novel}. Standard feature-based~\citep{alvarez2012feature} and trajectory replication-based techniques~\citep{chen2017integrating} are most commonly used in practice; appearance-based deep visual methods~\citep{ahmed2015improved,li2014deepreid} can also be used by recovery planner for person re-identification.

Additional planning and control procedures are required to incorporate the desired autonomous behaviors in emergency situations, \textit{e.g.}, when the recovery planner fails to re-identify the missing person, critically low battery or internal malfunctions detected, etc.  For instance, UAVs should be capable of making emergency landings and communicate its status to the person if possible. The UGVs and UAVs, on the other hand, can use some sort of emergency beacons, \textit{e.g.}, flashing lights, audio beeps, etc., for immediate attention.

\textbf{Social awareness in crowded area:} 
It is essential for person-following robots to maintain certain social rules while operating in a populated area~\citep{honig2018towards}. For instance, passing pedestrians on the right side, maintaining average human-walking speed, taking nearby persons' motions into account for planning, etc., are norms~\citep{dewantara2016generation,gockley2007natural} that a social robot should be aware of.
Therefore, application-specific social norms need to be modeled and translated into path planning and control rules in order to enable the desired behaviors. Enabling socially aware behaviors in mobile robots is an active research topic and researchers have been exploring these aspects for person-following robots in various environments    
such as airports~\citep{triebel2016spencer}, hospitals~\citep{ilias2014nurse}, and other crowded areas~\citep{ferrer2013robot}.

\textbf{Maintaining the norms of interaction:}
Lastly, the planning and control modules for person-following robots need to accommodate the norms of explicit and implicit human-robot interactions. In particular, various aspects such as the desired proximity behaviors, following angle~\citep{shanee2016influence}, turning and waiting behavior, etc., should be considered during trajectory planning. Additionally, application-specific choices such as whether to stay behind or side-by-side while following, desired speed, and relevant anticipative behaviors~\citep{mi2016system,granata2012framework} are essential considerations. Another important feature is to maintain the expected behavior during explicit interactions~\citep{islam2018understanding,hu2014design}, \textit{e.g.}, being stationary when the human is communicating, exhibiting correct acknowledgment responses, etc. These aspects of human-robot interaction are elaborately discussed in the following Section.

\subsection{Interaction}\label{Interaction}
Various forms of explicit and implicit interactions for person-following scenarios were discussed in Section~\ref{sec:cate}. The following discussion provides a summary of how these interactions take place, different methodologies used, and the related operational considerations.   

\subsubsection{Explicit Interaction}\label{explicitI}
Explicit interactions happen when there are direct communications between the human and the robot. Although most robots are equipped with peripheral devices and sometimes haptic sensors~\citep{ghosh2014following}, those are typically used for offline tasks (\textit{e.g.}, powering the robot, transferring program/data, sending emergency signals, etc.). On the other hand, communication paradigms based on speech, tags/markers, and hand gestures are used during operation for explicit human-robot interaction. 

\begin{figure}[t]
    \centering
    \begin{subfigure}[t]{0.225 \textwidth}
        \includegraphics[width=\linewidth]{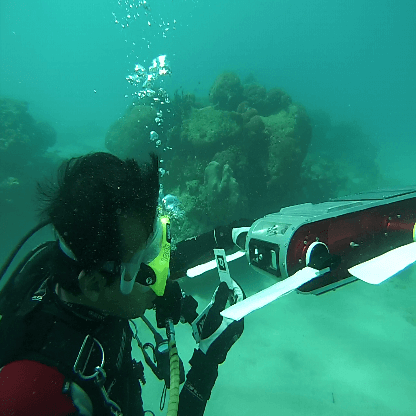} 
        \caption{Using AR-tags}
    \end{subfigure}
    \begin{subfigure}[t]{0.255\textwidth}
        \includegraphics[width=\linewidth]{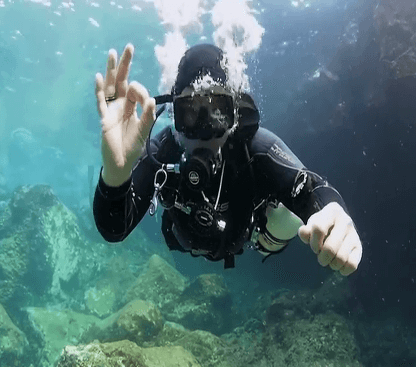}
        \caption{Using hand gestures}
    \end{subfigure}
    \vspace{-2mm}
    \caption{A diver is communicating instructions to an AUV during a mission~\citep{islam2018understanding}.}
    \label{fig:inter1}
\end{figure}

Verbal communication is convenient and commonly practiced in ground applications~\citep{sakagami2002intelligent}. Typically the person uses voice commands to convey simple instructions to the robot. The level of communication can vary from simple imperative commands (\textit{e.g.}, start/stop following, move left/right) to complex procedural instructions (\textit{e.g.}, a set of sequential tasks) depending on the application requirements. Systems for speech recognition and synthesis are very robust and commercially available these days. However, their usage is mostly limited to terrestrial applications.  

\begin{figure}[b]
    \centering
    \includegraphics[width=0.61\linewidth]{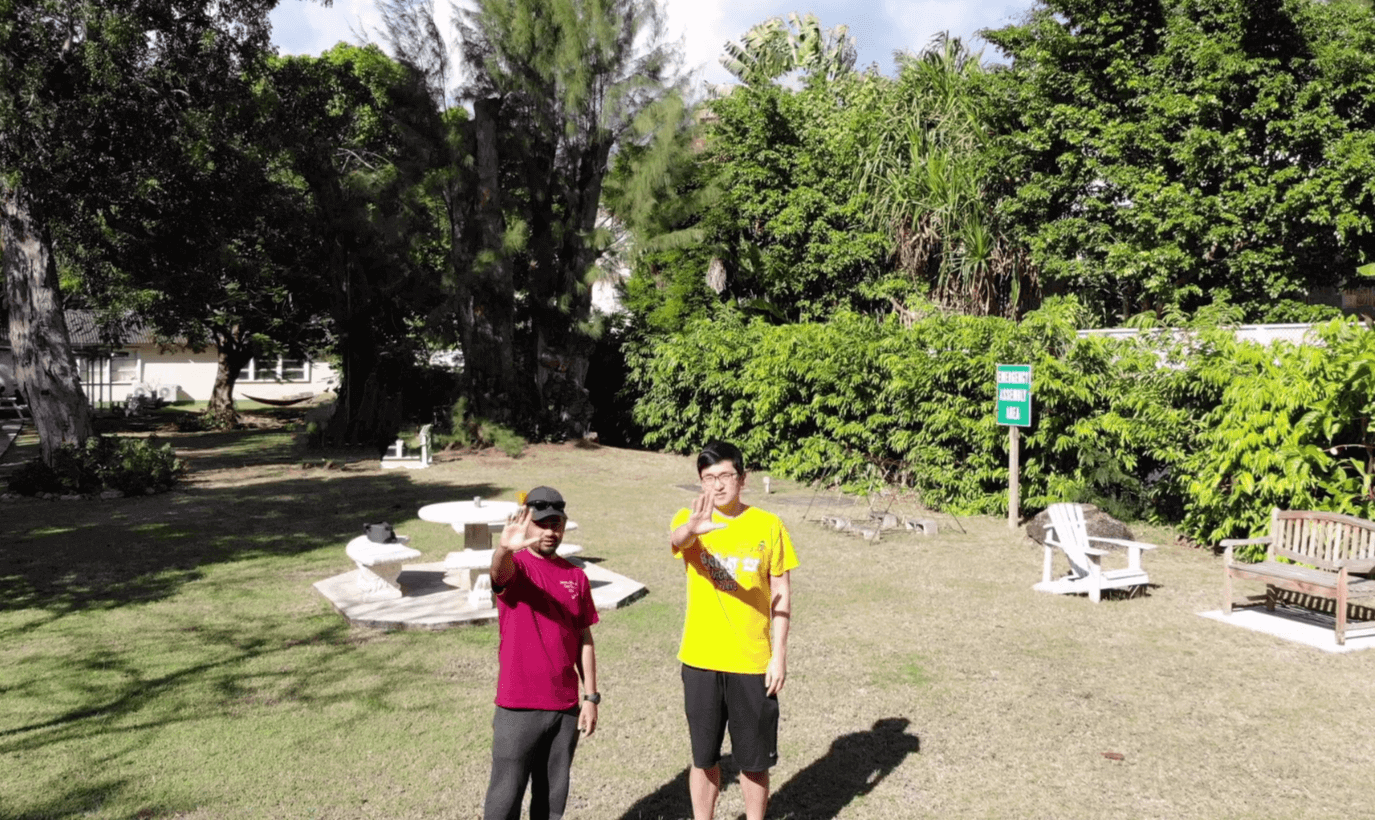}
    \includegraphics[width=0.375\linewidth]{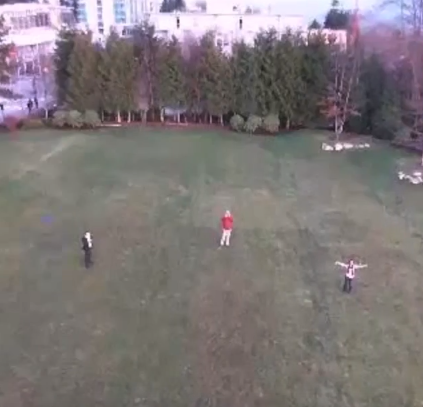}
    \vspace{-5mm}
    \caption{Illustrating the visual challenges of detecting hand gestures from a distant UAV \citep{monajjemi2016uav}: notice the minuscule appearance of humans on the right compared to the left image where the UAV is much closer to the person.}
    \label{fig:inter3}
\end{figure}

Tags or fiducial markers (\textit{e.g.}, ARTag, ARToolkit) have been used for underwater human-robot communication. Visual languages such as RoboChat~\citep{dudek2007visual} assign different sequences of symbolic patterns of those markers to a set of grammar rules (Figure~\ref{fig:inter1}). This is generally a robust way of communication because the fiducial markers are easy to detect in noisy underwater conditions. However, it is not very intuitive, and carrying a large set of markers during an underwater mission is inconvenient. Consequently, hand gesture-based communication paradigms~\citep{islam2017dynamic,chiarella2015gesture} are often preferred, where a sequence of hand gestures are used as symboling patterns instead of the tags. Detecting hand gestures in real-time is relatively more challenging than detecting markers; therefore, deep visual detectors are typically used for ensuring the robustness and accuracy of the system~\citep{islam2018understanding}.

An additional challenge for hand gesture-based communication in aerial applications is the relatively long and varying human-robot distances (Figure~\ref{fig:inter3}). Unlike in an underwater scenario, the person cannot come close to the robot and perform hand gestures in front of its camera. Consequently, the UAV might end up being too far to detect various kinds of hand gestures performed by the person~\citep{bruce2016tiny,monajjemi2016uav}. In such cases, it is often useful to use a reliable gesture (\textit{e.g.}, a static palm gesture, waving hands, etc.) to instruct the UAV to first come closer and then perform other hand gestures for communication~\citep{cauchard2015drone,naseer2013followme}. On the other hand, hand gesture-based communication is relatively less challenging in ground applications~\citep{alvarez2014gesture,marge2011comparing} and sometimes used even if a voice-based communication system is available. Moreover, it is often more feasible than voice-based communication in crowded environments~\citep{ferrer2013robot}, and in a multi-robot setting.

Smart devices and paired wearable devices are also commonly used for communicating human instructions, particularly by commercial UAVs~\citep{vasconcelos2016person,Skydio}, and interactive UGVs~\citep{faria2015probabilistic,burgard1998interactive}. Typically the humans use a set of menu options in order to instruct the robot to perform specific tasks. For instance, instructions such as start/stop recording videos, move to a particular direction, stop following, make an emergency landing, etc., are practically useful for interacting with person-following UAVs. On the other hand, a conversational user interface is needed for UGVs that serve as museum tour-guide robots, or as a personal assistant.

\subsubsection{Implicit Interaction}\label{implicitI}
The research studies on implicit interactions in person-following scenarios mostly concentrate on two aspects: the human perspective and the robot perspective. As mentioned in Section~\ref{sec:cate}, these aspects boil down to the following sets of mutual responsibilities in different stages of a person-following operation: 

\textit{\textbf{i. Spatial conduct}} consists of a set of desired proxemic behaviors~\citep{fleishman2018proxemic} of a robot while following a person in an HRI setting. This behavioral quantification is important to define safe person-following trajectories and to model proximity control parameters~\citep{yamaoka2008close} for following and waiting, while engaging in, and during an explicit interaction.  

\textit{\textbf{ii. Appearance and gaze conduct}} consists of desired responsive behaviors during human-robot communication~\citep{zender2007human} and non-responsive gaze behaviors during following, approaching and handing over scenarios~\citep{moon2014meet}, etc. Moreover, for person-following robots that stay in-front~\citep{nikdel2018the,jung2012control}, the companion human's gaze behavior is an important feature to track in order to predict their motion. Additionally, it helps the robot to identify when the person is about to start an explicit interaction and to plan accordingly, \textit{e.g.}, slow down or stop, prepare to detect and interpret the communicated instructions, etc.

\textit{\textbf{iii. Motion conduct}} refers to a set of desired person-following motion trajectories of the robot in different situations. It includes motion models for following a person from different directions (\textit{e.g.}, from behind, side-by-side, at an angle, etc.), turning behaviors, and waiting behaviors. Additionally, the expected motion behavior of the robot when its human companion is interacting with other people or goes out of its sight$-$ are important design considerations~\citep{granata2012framework,gockley2007natural}. Motion conduct rules are used by the planning component of the system in order to maintain the desired motion behaviors. Therefore, prior knowledge about human motion (\textit{e.g.}, walking, swimming, etc.) and the overall interaction model can facilitate the design of those anticipating motion behaviors~\citep{hu2014design} for person-following robots. 

The modalities and characteristics of implicit interactions are difficult to quantify in terms of technicality. It calls for rigorous user studies and feasibility analysis to formulate the right research questions and their effective solutions.

%%%%% Table 6
\begin{table*}[ht]
\centering
\footnotesize
\caption{An ordered collection of the person-following systems that are discussed in this paper; they are mentioned in a reverse chronological order and grouped based on their primary focus (\textit{e.g.}, perception, planning, control, and interaction).
[note that the diamond signs ($^\Diamond$) indicate that the corresponding techniques are not specifically about person-following robots, yet they are applicable and/or relevant] }
\begin{tabular}{|p{1.45cm}||p{6cm}|p{3.5cm}|p{4.5cm}|}
\hline
& Perception & Planning and control & Interaction \\
\hline \hline

%%%%%%%%%%%%%%%%%%%%%%%%%%%%%%%%%%%%%%%%%%%%%%%%%%%%%%%%%%%%
Ground ($2010$-$18$) & 
\cite{popov2018detection,wang2018accurate,chi2018gait,jiang2018classification,gupta2017novel,chen2017integrating};
\cite{chen2017decentralized}$^\Diamond$; 
\cite{chen2017person,wang2017real,do2017reliable}; 
\cite{cao2017realtime}$^\Diamond$; 
\cite{guevara2016vision,koide2016identification}; 
\cite{faria2015probabilistic}$^\Diamond$; 
\cite{leigh2015person,babaians2015skeleton,eisenbach2015user,isobe2014human,cai2014human,ilias2014nurse,pairo2013person};
\cite{pradhan2013indoor}$^\Diamond$; 
\cite{munaro2013software,satake2013visual,cao2013specific,susperregi2013rgb,basso2013fast,awai2013hog,yoon2013depth,satake2012sift,chung2012detection,alvarez2012feature,gascuena2011agent};
\cite{dollar2010fastest}$^\Diamond$; 
\cite{germa2010vision,brookshire2010person}
&
\cite{nikdel2018the,pierre2018end,wang2018person,chen2018folo,huskic2017outdoor};
\cite{chen2017socially}$^\Diamond$; 
\cite{masuzawa2017development,mi2016system,peng2016tracking};
\cite{cosgun2016anticipatory}$^\Diamond$;
\cite{sung2016hierarchical};
%\cite{ghosh2014following}$^\Diamond$; 
\cite{digiacomcantonio2014self,tarokh2014vision,park2013autonomous};
\cite{pradhan2013indoor}$^\Diamond$; 
\cite{jung2012control,morioka2012control,doisy2012adaptive};
\cite{yamaoka2010model,tarokh2010vision}
&
\cite{honig2018towards}; 
\cite{ahn2018formation}$^\Diamond$; 
\cite{fleishman2018proxemic};
\cite{pourmehr2017robust}$^\Diamond$; 
\cite{triebel2016spencer,alves2016study,shanee2016influence}; 
\cite{dewantara2016generation}$^\Diamond$; 
\cite{thomason2015learning}$^\Diamond$; 
\cite{alvarez2014gesture,hu2014design,budgee}; 
\cite{cifuentes2014human}$^\Diamond$; 
\cite{moon2014meet}$^\Diamond$; 
\cite{doisy2013spatially,cosgun2013autonomous}; 
\cite{ferrer2013robot}$^\Diamond$; 
\cite{granata2012framework,marge2011comparing}
\\
\hline

Ground ($2000$-$09$) & 
\cite{hu2009reliable,hu2007robust};  
\cite{dollar2009integral}$^\Diamond$;
\cite{bajracharya2009fast}$^\Diamond$;
\cite{germa2009vision,satake2009robust,shaker2008fuzzy,liem2008hybrid,handa2008person,zender2007human,takemura2007person,calisi2007person,chen2007person,itoh2006development,yoshimi2006development,kobilarov2006people}; 
\cite{zhu2006fast}$^\Diamond$; 
\cite{kwon2005person,kwolek2004person}; 
\cite{sedighi2004autonomous}$^\Diamond$; 
\cite{hirai2003visual}
&  
\cite{satake2009robust,luo2009human};
\cite{muller2008socially}$^\Diamond$;
\cite{hoeller2007accompanying,chivilo2004follow,tarokh2003case}
&
\cite{yamaoka2008close}$^\Diamond$; 
\cite{kuno2007museum,gockley2007natural}; 
\cite{syrdal2007personalized}$^\Diamond$;
\cite{huttenrauch2006investigating}$^\Diamond$;
\cite{yoshikawa2006responsive}$^\Diamond$;
\cite{matsumaru2005mobile,kulykukin2004human}
\cite{sakagami2002intelligent}$^\Diamond$;

\\
\hline

Ground (*-$1999$) & 
\cite{stauffer1999adaptive}$^\Diamond$; 
\cite{schlegel1998vision,darrell1998integrated};   
\cite{papageorgiou1998general}$^\Diamond$; 
\cite{yamane1998person}$^\Diamond$;
\cite{wren1997pfinder}$^\Diamond$; 
\cite{azarbayejani1996real}$^\Diamond$;
&  
\cite{sidenbladh1999person}; 
\cite{stentz1994optimal}$^\Diamond$;
\cite{espiau1992new}$^\Diamond$;  
& 
\cite{piaggio1998optical};
\cite{burgard1999experiences}$^\Diamond$;
\cite{burgard1998interactive}$^\Diamond$; 
\\
\hline \hline

%%%%%%%%%%%%%%%%%%%%%%%%%%%%%%%%%%%%%%%%%%%%%%%%
Underwater ($2010$-$18$) & 
\cite{islam2018towards,islam2017mixed};
\cite{mandic2016underwater,hari2015robust,miskovic2015tracking,gemba2014partial,demarco2013sonar} 
&  
\cite{islam2018towards}; 
\cite{zadeh2016auv}$^\Diamond$; 
\cite{shkurti2017underwater}$^\Diamond$; 
\cite{meger20143d}$^\Diamond$
\cite{janabi2011comparison}$^\Diamond$
&  
\cite{islam2018understanding,islam2017dynamic,gomez2019caddy,chiarella2018novel}; 
\cite{fulton2018robot}$^\Diamond$;
\cite{chiarella2015gesture}$^\Diamond$;
\cite{stilinovic2015auv} 
%\cite{wu2015towards}$^\Diamond$;
\\
\hline

Underwater (*-$2009$) & 
\cite{lennartsson2009electric,sattar2009underwater}; 
\cite{sattar2006performance}$^\Diamond$
&  
\cite{sattar2009robust}; 
\cite{corke2007experiments}$^\Diamond$; 
\cite{rosenblatt2002behavior}$^\Diamond$
& 
\cite{sattar2009underwater};
\cite{xu2008natural}$^\Diamond$; 
\cite{dudek2007visual}$^\Diamond$
\\
\hline \hline

%%%%%%%%%%%%%%%%%%%%%%%%%%%%%%%%%%%%%%%%%%%%%%%%
Aerial & 
\cite{Skydio,mueller2016persistent,vasconcelos2016person}; 
\cite{chakrabarty2016autonomous}$^\Diamond$; 
\cite{bartak2015any}$^\Diamond$; 
\cite{de2015board,portmann2014people,pestana2014computer,naseer2013followme,graether2012joggobot,higuchi2011flying};
\cite{kumar2011visual}$^\Diamond$; 
\cite{teuliere2011chasing}$^\Diamond$; 
\cite{gaszczak2011real}$^\Diamond$ 
&  
\cite{Skydio,Staaker}; 
\cite{huh2013integrated}$^\Diamond$; 
\cite{lugo2014framework}$^\Diamond$; 
\cite{gong2011multi}$^\Diamond$; 
\cite{tomic2012toward}$^\Diamond$;
\cite{teuliere2011chasing}$^\Diamond$;
\cite{kim2008real}$^\Diamond$
&  
\cite{nagyflying,bruce2016tiny,vasconcelos2016person,monajjemi2016uav,cauchard2015drone}; 
\cite{gioioso2014flying}$^\Diamond$;
\cite{lichtenstern2012prototyping}$^\Diamond$;
\cite{tisdale2009autonomous}$^\Diamond$; 
\cite{mezouar2002path};$^\Diamond$ 
\\
\hline

%%%%%%%%%%%%%%%%%%%%
\end{tabular}
\label{all_sum}
\end{table*}

%%%%%%%%%%%%%%%%%%%%%%%%%%%%%%%%%%%%%%%%%%%%%%%%%%%%%%%
%%%%% Table 7
\begin{table*}[t]
\centering
\footnotesize
\caption{A set of qualitative comparisons is presented among a number of prominent person-following systems reported over the last decade ($2009$-$2019$). They are compared based on a subset of these items$-$ 
(i) Detection \& Tracking: qualitative performance, 
(ii) Online (ReID): whether online learning (person re-identification or recovery) module is used, 
(iii) Optimal Planning/Control: optimality of the underlying planning and control modules, 
(iv) Obst. Avd.: presence of \textit{obstacle avoidance} feature (for UGVs/UAVs), 
(v) Exp. (Imp.): availability of some forms of explicit (implicit) interaction, 
(vi) Interactive: availability of interactive user interfaces (for UGVs/UAVs),
(vii) Multi-H: applicability of the system for multiple human-following, 
(viii) Outdoors: applicability in outdoors (for UGVs/UAVs), 
(ix) Socially Aware: availability of socially-compliant planning/interaction modules (for UGVs),  
(x) Crowded Places: applicability in crowded/populated area (for UGVs/UAVs), 
(xi) Invariance to: whether the tracking performance is invariant to divers' \textit{appearance, motion, and wearables} (of AUV/ASVs), 
(xii) Coastal Waters: applicability in coastal and shallow waters (for AUVs/ASVs),
(xiii) Visibility: applicability in \textit{poor/no} visibility conditions (for AUVs/ASVs), and 
(xiv) GPS-denied: applicability in GPS-denied environments (for UAVs).}

\vspace{2mm}
\begin{tabular}{|m{2.6cm}|m{1.3cm}|m{1cm}|m{1.7cm}|m{0.6cm}|m{0.7cm}|m{0.8cm}|m{1cm}|m{0.95cm}|m{0.8cm}|m{0.9cm}|}
\multicolumn{11}{l}{(a) Person-following systems for UGVs} \\ \hline
 & \multicolumn{4}{c|}{Perception, Planning, \& Control} & \multicolumn{2}{c|}{Interaction} & \multicolumn{4}{c|}{Multi-H Support \& General Applicability} \\ \cline{2-11}
 & Detection \& Tracking & Online (ReID)  & Optimal Plan- ning/Control & Obst. Avd. & Exp. (Imp.) & Inter- active & Multi-H & Outdoors & Socially Aware & Crowded Places \\
\hline \hline

\cite{wang2018accurate}   & \checkmark \checkmark \checkmark& $\times$($\times$) & $\times$/$\times$ & $\times$ & $\times$($\times$)& $\times$& $\times$ & \checkmark& $\times$ & \checkmark \\ \hline
%obstacle:maybe

\cite{nikdel2018the}& \checkmark \checkmark &  $\times$ ($\times$) & \checkmark/\checkmark & \checkmark & \checkmark(\checkmark) & $\times$& $\times$ & $\times$ & $\times$ & $\times$\\ \hline

\cite{chen2018folo}  & \checkmark \checkmark \checkmark &  $\times$($\times$) & $\times$/$\checkmark$ & $\times$ & $\times$($\times$) & $\times$ & $\times$ & $\times$ & $\times$ & \checkmark\\ \hline

\cite{chen2017person} & 
\checkmark \checkmark \checkmark & \checkmark($\times$) & $\times$/\checkmark  & $\times$ & $\times$($\times$) & $\times$ & $\times$ & \checkmark & $\times$ & $\times$ \\ \hline

\cite{gupta2017novel} & \checkmark \checkmark \checkmark & \checkmark (\checkmark)
& $\times$/\checkmark & $\times$ & $\times$(\checkmark) & $\times$ & $\times$ & \checkmark & $\times$ & \checkmark\\ \hline

\cite{chen2017integrating} & \checkmark \checkmark \checkmark & \checkmark (\checkmark) & $\times$/$\times$ &$\times$ & $\times$(\checkmark)& $\times$& $\times$ & \checkmark& $\times$ & \checkmark \\ \hline

%\cite{chen2017socially} & \checkmark \checkmark &  \checkmark($\times$) & \checkmark/\checkmark & \checkmark & $\times$(\checkmark) & $\times$& \checkmark & $\times$ & \checkmark & \checkmark\\ \hline
\cite{huskic2017outdoor} & \checkmark \checkmark &  $\times$($\times$) & \checkmark/\checkmark & \checkmark & $\times$(\checkmark) & $\times$ & $\times$ & \checkmark & $\times$ & \checkmark\\ \hline

\cite{koide2016identification} & \checkmark \checkmark \checkmark & \checkmark (\checkmark)&$\times$/$\times$&$\times$ & $\times$(\checkmark) & $\times$& $\times$ & \checkmark & $\times$ & \checkmark\\ \hline

\cite{triebel2016spencer}  & \checkmark \checkmark \checkmark &  \checkmark($\times$) & \checkmark/$\times$ & \checkmark & \checkmark(\checkmark) & \checkmark& \checkmark & $\times$ & \checkmark & \checkmark\\ \hline

\cite{sung2016hierarchical} & \checkmark \checkmark \checkmark & $\times$($\times$) & $\times$/$\times$ & $\times$ & $\times$(\checkmark) & $\times$& $\times$ & $\times$ & $\times$ & \checkmark\\ \hline

%\cite{chen2017decentralized} & \checkmark \checkmark&\checkmark ($\times$) & \checkmark & $\times$ & \checkmark & & & & $\times$ & $\times$ & $\times$\\ \hline %% no following component
%\cite{cosgun2016anticipatory}  & \checkmark \checkmark &  $\times$($\times$) & \checkmark/$\times$ & \checkmark & $\times$(\checkmark) & $\times$& \checkmark & $\times$ & \checkmark & \checkmark\\ \hline  %this is not following paper. it's a planning paper.

\cite{leigh2015person} & \checkmark \checkmark & $\times$ ($\times$)& $\times$/\checkmark &$\times$ & $\times$(\checkmark) & $\times$& \checkmark & \checkmark & $\times$ & $\times$\\ \hline

\cite{eisenbach2015user}   & \checkmark \checkmark \checkmark& $\times$(\checkmark) & $\times$/$\times$ & \checkmark & $\times$(\checkmark)& $\times$& $\times$ & $\times$& $\times$& \checkmark\\ \hline

\cite{hu2014design}& \checkmark \checkmark &  $\times$($\times$) & \checkmark/\checkmark & \checkmark & $\times$(\checkmark) & $\times$& $\times$ & $\times$ & \checkmark & $\times$\\ \hline

\cite{munaro2013software} & \checkmark \checkmark \checkmark &  \checkmark(\checkmark) & $\times$/$\times$ & \checkmark & $\times$(\checkmark) & $\times$ & \checkmark & $\times$ & $\times$ & \checkmark\\ \hline

\cite{park2013autonomous}  & \checkmark \checkmark &  \checkmark($\times$) & \checkmark/$\times$ & \checkmark & $\times$(\checkmark) & $\times$ & $\times$ & \checkmark & \checkmark & \checkmark\\ \hline 

\cite{cosgun2013autonomous}  & \checkmark \checkmark &  $\times$($\times$) & \checkmark/$\times$ & \checkmark & \checkmark(\checkmark) & \checkmark& $\times$ & $\times$ & \checkmark & $\times$\\ \hline

\cite{chung2012detection}& \checkmark \checkmark \checkmark &  $\times$(\checkmark) & $\times$/$\times$ & $\times$ & $\times$(\checkmark) & $\times$& $\times$ & $\times$ & $\times$ & \checkmark\\ \hline

\cite{doisy2012adaptive} & \checkmark \checkmark &  $\times$(\checkmark) & \checkmark/\checkmark & \checkmark & $\times$(\checkmark) & $\times$& $\times$ & $\times$ & $\times$ & $\times$\\ \hline

\cite{granata2012framework} & \checkmark \checkmark &  $\times$($\times$) & \checkmark/\checkmark & \checkmark & \checkmark(\checkmark) & \checkmark& $\times$ & $\times$ & \checkmark & \checkmark\\ \hline

\cite{germa2010vision} & \checkmark \checkmark \checkmark &  $\times$(\checkmark) & $\times$/\checkmark & $\times$ & $\times$(\checkmark) & $\times$& $\times$ & $\times$ & $\times$ & \checkmark\\ \hline

%% \checkmark
%% $\times$
\end{tabular}
%%%%%%%%%%%%%%%%%%%%
\vspace{2mm}

\begin{tabular}{|m{2.9cm}|m{1.3cm}|m{4.2cm}|m{1.6cm}|m{1.1cm}|m{0.95cm}|m{1cm}|m{0.7cm}|}
\multicolumn{8}{l}{(b) Person-following systems for AUVs [systems for ASVs are marked with a star ($*$) sign]} \\ \hline
 & \multicolumn{3}{c|}{Perception, Planning, \& Control} & Interaction & \multicolumn{3}{c|}{Multi-H Support \& Feasibility} \\ \cline{2-8}
 & Detection \& Tracking  & Invariance to: $<$Appearance, Motion, Wearables$>$ $|$ Online (ReID) & Optimal Plan- ning/Control & H-to-R (R-to-H) & Multi-H & Visibility Poor/No & Coastal Waters  \\
\hline  \hline 

\cite{islam2018towards} & \checkmark \checkmark \checkmark   & $<$\checkmark, \checkmark, \checkmark$>$ $|$  $\times$ ($\times$)  & \checkmark $/$ \checkmark & \checkmark($\times$) & \checkmark  & \checkmark/$\times$ & \checkmark \\ \hline

\cite{islam2018understanding} & \checkmark \checkmark \checkmark   & $<$\checkmark, \checkmark, \checkmark$>$ $|$ $\times$ ($\times$) & \checkmark $/$ \checkmark & 
$\times$($\times$) & $\times$  & \checkmark/$\times$ & \checkmark \\ \hline

\cite{islam2017mixed} & \checkmark \checkmark  & $<$\checkmark, $\times$, \checkmark$>$ $|$  $\times$ ($\times$)  & $\times/\times$ & 
$\times$($\times$) & $\times$  & $\times$/$\times$ & \checkmark \\ \hline

\cite{mandic2016underwater}  & \checkmark \checkmark \checkmark  & $<$\checkmark,  \checkmark , \checkmark$>$ $|$  $\times$ ($\times$)  & $\times/$\checkmark  & 
$\times$($\times$) & $\times$  & \checkmark/\checkmark & $\times$ \\ \hline

\cite{hari2015robust} & \checkmark \checkmark   & $<$\checkmark,  \checkmark , \checkmark$>$ $|$  $\times$ ($\times$)  & $\times/\times$ & 
$\times$($\times$) & $\times$  & \checkmark/\checkmark & \checkmark \\ \hline

\cite{miskovic2015tracking}$^*$ & \checkmark \checkmark \checkmark   & $<$\checkmark,  \checkmark , \checkmark$>$ $|$  $\times$ ( \checkmark)  &  \checkmark$/$\checkmark & 
$\times$(\checkmark) &  \checkmark  & \checkmark/\checkmark & $\times$ \\ \hline

\cite{gemba2014partial} & \checkmark \checkmark   & $<$\checkmark,  \checkmark , \checkmark$>$ $|$  $\times$ ($\times$)  & $\times/\times$ & 
$\times$($\times$) & $\times$  & \checkmark/\checkmark & \checkmark \\ \hline

\cite{demarco2013sonar} & \checkmark \checkmark  & $<$\checkmark,  \checkmark , \checkmark$>$ $|$  $\times$ ($\times$)  & $\times/\times$  & 
$\times$($\times$) & $\times$  & \checkmark/\checkmark & \checkmark \\ \hline

\cite{sattar2009underwater} & \checkmark \checkmark  & $<$\checkmark, $\times$, $\times$$>$ $|$  $\times$ ($\times$)  & $\times$/\checkmark & 
$\times$($\times$) & $\times$  & $\times$/$\times$ & \checkmark \\ \hline

\end{tabular}
%%%%%%%%%%%%%%%%%%%%
\vspace{2mm}

\begin{tabular}{|m{2.6cm}|m{1.4cm}|m{0.9cm}|m{1.7cm}|m{0.7cm}|m{0.7cm}|m{0.8cm}|m{0.95cm}|m{0.95cm}|m{0.8cm}|m{0.95cm}|}
\multicolumn{10}{l}{(c) Person-following systems for UAVs [commercially available UAVs are marked with a star ($*$) sign]} \\ \hline
 & \multicolumn{4}{c|}{Perception, Planning, \& Control} & \multicolumn{2}{c|}{Interaction} & \multicolumn{4}{c|}{Multi-H Support \& General Applicability} \\ \cline{2-11}
 & Detection \& Tracking & Online (ReID)  & Optimal Plan- ning/Control & Obst. Avd. & Exp. (Imp.) & Inter- active & Multi-H & Outdoors & GPS- denied & Crowded Places  \\
\hline \hline

\cite{Skydio}$^*$ & 
\checkmark \checkmark \checkmark & $\times$(\checkmark) & \checkmark$/$\checkmark 
 & \checkmark & \checkmark($\times$) & \checkmark 
 & \checkmark & \checkmark & $\times$ & \checkmark \\ \hline
 
\cite{vasconcelos2016person} & 
\checkmark \checkmark & $\times$(\checkmark) & $\times/\times$ 
 & $\times$ & \checkmark($\times$) & $\times$ 
 & $\times$ & \checkmark & $\times$ & $\times$ \\ \hline
 
\cite{mueller2016persistent} & 
\checkmark \checkmark \checkmark & $\times$(\checkmark) & $\times/$\checkmark 
 & $\times$ & $\times$($\times$) & \checkmark 
 & $\times$ & \checkmark & $\times$ & \checkmark \\ \hline
 
\cite{de2015board} & 
\checkmark \checkmark \checkmark & $\times$($\times$) & $\times/$\checkmark 
 & $\times$ & $\times$($\times$) & $\times$ 
 & $\times$ & \checkmark & $\times$ & $\times$ \\ \hline
 
\cite{portmann2014people} & 
\checkmark \checkmark \checkmark & $\times$($\times$) & $\times/\times$ 
 & $\times$ & $\times$($\times$) & $\times$ 
 & $\times$ & \checkmark & $\times$ & \checkmark \\ \hline
 
 \cite{pestana2014computer} & 
\checkmark \checkmark & $\times$(\checkmark) & $\times/$\checkmark 
 & $\times$ & $\times$($\times$) & $\times$ 
 & $\times$ & \checkmark & \checkmark & \checkmark \\ \hline
 
\cite{naseer2013followme} & 
\checkmark \checkmark & $\times$($\times$) & $\times/$\checkmark 
 & $\times$ & \checkmark($\times$) & $\times$ 
 & $\times$ & $\times$ & \checkmark & \checkmark \\ \hline
 
\cite{Staaker}$^*$ & 
\checkmark \checkmark \checkmark & $\times$(\checkmark) & \checkmark$/$\checkmark 
 & $\times$ & \checkmark($\times$) & $\times$ 
 & $\times$ & \checkmark & $\times$  & \checkmark \\ \hline
 
 \cite{higuchi2011flying} & 
\checkmark \checkmark & $\times$($\times$) & $\times/$\checkmark 
 & $\times$ & \checkmark($\times$) & $\times$ 
 & $\times$ & \checkmark & $\times$  & \checkmark \\ \hline

\end{tabular}
%%%%%%%%%%%%%%%%%%%%

\label{sel_sum}
\end{table*}

\section{Qualitative Analysis: Feasibility, Practicality, and Design Choices}\label{sec:critic}
An overwhelming amount of research work and industrial contributions have enriched the literature on autonomous person-following. This paper highlights and organizes these into a categorical study; to further assist the readers to navigate through the large body of literature, they are presented in an ordered and organized fashion in Table~\ref{all_sum}. 
This Section analyzes a number of prominent person-following systems and provides a comparative discussion in qualitative terms. A summary of this qualitative analysis is depicted in Table~\ref{sel_sum}.

\subsection{Detection and Tracking Performance}
An important consideration for designing a perception module is the desired level of detection accuracy and tracking performance given the operating constraints. This impacts the choices of sensors and on-board computational hardware as well. For instance, person-following UGVs can accommodate multiple sensors, \textit{e.g.}, combinations of cameras, sonars, laser scanners or RGBD cameras, etc. 
Therefore, it is generally a good practice to adopt sensor fusion schemes~\citep{wang2018accurate,nikdel2018the,susperregi2013rgb} for ensuring accurate feature-based detection and tracking at a fast rate. 
If only a single exteroceptive sensor is (can be) used, it requires more sophisticated techniques such as deep visual models, online learning-based models, etc., to ensure reliable perception~\citep{wang2018person,chen2017person}; these models are computationally demanding and typically require single-board supercomputers~\citep{Jetson} for real-time inference. However, if there are constraints on power, the use of UWB/RFID tags~\citep{laneurit2016trackbod,germa2010vision} is ideal for designing effective low-power solutions.

The constraints on power consumption and resource utilization are more important considerations for person-following AUVs and UAVs. Hence, using domain-specific prior knowledge such as modeling diver's swimming pattern by AUVs~\citep{islam2017mixed}, perspective filtering by UAVs~\citep{de2015board}, etc., can facilitate the design of efficient trackers. Nevertheless, on-board supercomputers~\citep{Jetson} and/or edge devices~\citep{coral} can be used to run deep visual trackers in real-time~\citep{islam2018towards,Skydio}. Moreover, paired connectivity with the companion human, \textit{e.g.}, paired GPS receiver by UAVs~\citep{Staaker}, acoustic link by ASVs~\citep{miskovic2015tracking}, can provide reliable and fast tracking performances at a low power budget.

The methodological details of the above-mentioned perception modules are already discussed in Section~\ref{sec:approach_Perception}. A qualitative comparison among them is provided in the first column of Table~\ref{sel_sum}. The comparison also includes two other important features, \textit{i.e.}, whether online learning is used~\citep{gupta2017novel,park2013autonomous}, and whether person re-identification/recovery is considered~\citep{chen2017integrating,doisy2012adaptive}. Additionally, for diver-following systems, the invariance to divers' appearance, motion, and wearables are taken into account for comparison.
While interpreting this qualitative comparison, it is to be noted that multiple check-mark (\checkmark) symbols in the first comparison (\textit{i.e.}, for detection and tracking) represent the quality of a proposed solution on a scale of one to three, where three means state-of-the-art performance. In all other columns of Table~\ref{sel_sum}, the check-mark (\checkmark) and cross ($\times$) symbols independently represent \textit{yes}, and \textit{no}, respectively, for their corresponding comparisons.

\subsection{Optimal Planning and Control}
A few application-specific requirements, particularly the degree of autonomy and the presence of dynamic agents/obstacles in the operating environment directly influence the design choices for planning and control modules of a person-following robot. 
For instance, in predominately static settings, robots can rely on their human companion for collision-free navigation, \textit{i.e.}, plan to maintain a constant distance while assuming there will be no interfering agents along the way. This approach, often with additional features for obstacle-avoidance, is feasible in underwater scenarios~\citep{islam2018understanding}, and adopted in many ground applications~\citep{sung2016hierarchical,koide2016identification} of person-following. 
However, as discussed in Section~\ref{sec:plan_ctr}, optimal planning with the consideration of dynamic obstacles, motion and interaction from other humans, norms of social/public places, etc.,  are essential for robots operating in 
crowded area~\citep{park2013autonomous,granata2012framework}, social settings~\citep{triebel2016spencer,cosgun2013autonomous}, and in challenging outdoor scenarios~\citep{mueller2016persistent,Staaker}.

On the other hand, complex motion planning requires dense knowledge about the environment, which impacts the choice and modality of sensors. For instance, UGVs operating in known indoor environments can take advantage of a global map~\citep{nikdel2018the} in order to accurately plan to navigate while avoiding obstacles~\citep{triebel2016spencer}. Even when global map is not available, 3D sensing capabilities (\textit{e.g.}, camera with sonar/LRF/IR sensor, or with multiple cameras) are needed to get a localized 3D information about the world, which can be used for SLAM-based navigation~\citep{Skydio,huskic2017outdoor}. Furthermore, based on application-specific requirements, the rules of social norms and desired implicit behaviors of the robot need to be modeled as prior knowledge and eventually incorporated into the planning and control modules. These aspects are also considered for the qualitative comparison illustrated in Table~\ref{sel_sum}.

\subsection{Interactivity and General Feasibility}
A number of important design choices depend on the desired level of interactivity between a robot and its companion human~\citep{cosgun2013autonomous,granata2012framework}. This influences the choice of sensors or peripheral devices (\textit{e.g.}, interactive screen, voice interface, paired application, etc.), and the design of important perception and planning modules (\textit{e.g.}, hand-gesture recognition, action recognition, planning for implicit interaction, etc.). Additionally, some aspects such as whether multiple-human support (\textit{i.e.}, following as a group) or social awareness is needed, following ahead or behind, etc., are essential considerations while designing a person-following system. 
These interactivity requirements need to be formulated by thorough user experiments for practical applications~\citep{triebel2016spencer,gockley2007natural}.

Several features pertaining to the interactivity and general feasibility for person-following robots are considered for qualitative comparison in Table~\ref{sel_sum}. These aspects, relevant design issues based on various use-cases, and the corresponding state-of-the-art solutions  for ground, underwater, and aerial scenarios are elaborately discussed in this paper. As evident from these discussions, the vast majority of the literature on person-following robots address various research problems in ground scenarios. It is safe to say that the current state-of-the-art systems provide very good solutions to these problems. However, the social and behavioral aspects of these systems require more attention from the research community. In addition, better and smarter methodologies are required to address the unique challenges of underwater and aerial scenarios. These aspects, and other important research directions are highlighted in the following Section.

\section{Prospective Research Directions}\label{sec:future}
The following subsections discuss a number of active research areas and open problems that are naturally challenging and are potentially useful in person-following applications.

\subsection{Following a Team}
\begin{figure}[h]
    \centering
    \begin{subfigure}[t]{0.205 \textwidth}
        \includegraphics[width=\linewidth]{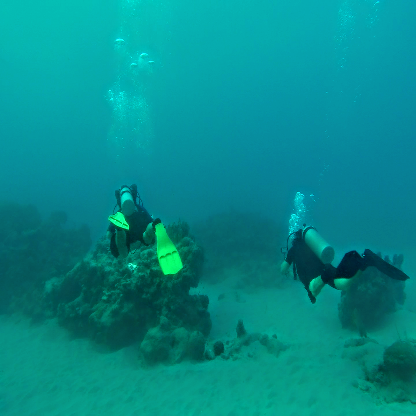} 
        \caption{Underwater scenario}
    \end{subfigure}
    \begin{subfigure}[t]{0.274\textwidth}
        \includegraphics[width=\linewidth]{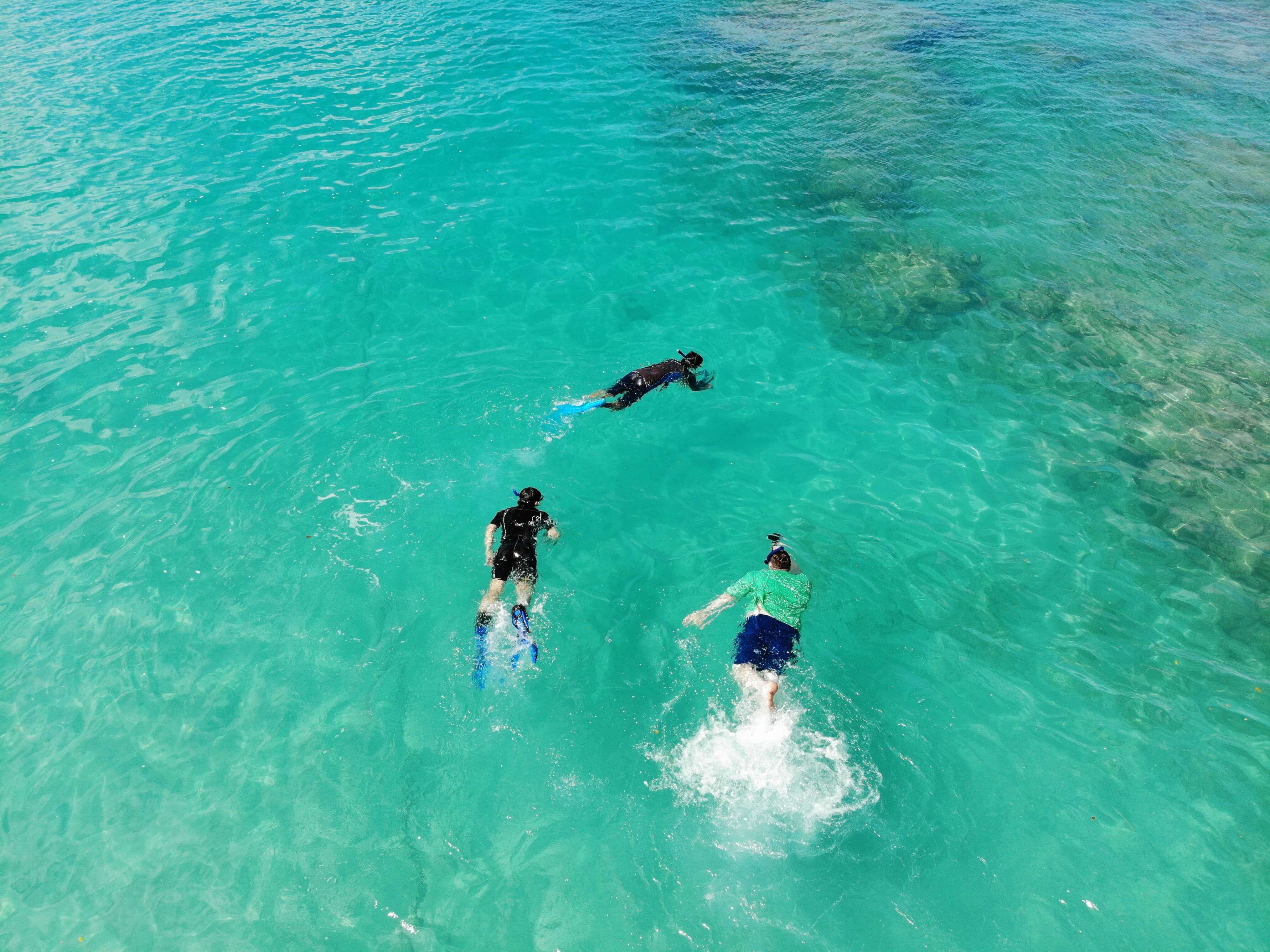}
        \caption{Aerial scenario}
    \end{subfigure}
    \vspace{-2mm}
    \caption{{  Views from robots' camera while following a team of people.}}
    \label{fig:inter4}
    \end{figure}

Many underwater missions consist of multiple divers who work as a team during the deployment (Figure~\ref{fig:inter4}). Following the team as a whole is operationally more efficient in general. Similar scenarios arise when filming a social or sports event using UAVs. The perception problem can be easily solved by a simple extension (\textit{i.e.}, by allowing the detection of multiple humans); however, motion planning and control modules are not straight-forward. Moreover, the rules for spatial conduct and interaction need to be identified and quantified. Tracking a team of independently moving objects is a challenging problem in general~\citep{shu2012part}; it gets even more challenging in a 3D environment while dealing with real-time constraints. Despite the challenges, it is potentially invaluable in numerous applications of person-following robots~\citep{phantom,shkurti2012multi}.

\subsection{Following as a Team (Convoying)}
Multi-robot convoys, being led by a human, is useful in cooperative estimation problems~\citep{rekleitis2001multi}.  
A simple approach to this problem is to assign leader-follower pairs; that is, one of the robots is assigned to follow the person, and every other robot is individually assigned another robot as its leader. Each robot follows its leader and together they move as a team. Another approach is to let the robots communicate with each other and cooperatively plan their motions. The underlying planning pipeline is similar to that of a multi-robot convoying problem, which is particularly challenging in underwater and aerial scenarios~\citep{shkurti2017underwater,minaeian2016vision}. Moreover, this can be further generalized into the problem of \textit{following a group of people by a team of autonomous robots} in a cooperative setting. However, it requires a complex cooperative planning pipeline to achieve the optimal positioning and motion trajectories for each robot, which is an open problem as well.

\subsection{Following Behind or Ahead?} 
There are scenarios where it is ideal to have the robot stay ahead of the person while following. Hands-free shopping cart robots, for instance, should stay ahead of the human, not behind~\citep{nikdel2018the,kuno2007museum}. Another prime example is the person-following UAVs that record sports activities; they should be able to move around and take snapshots from different directions to get the best perspective~\citep{Skydio}. Therefore, the traditional systems and methodologies for \textit{following from behind} are not very useful in these applications.

   \begin{figure}[t]
    \centering
    \begin{subfigure}[t]{0.275 \textwidth}
        \includegraphics[width=\linewidth]{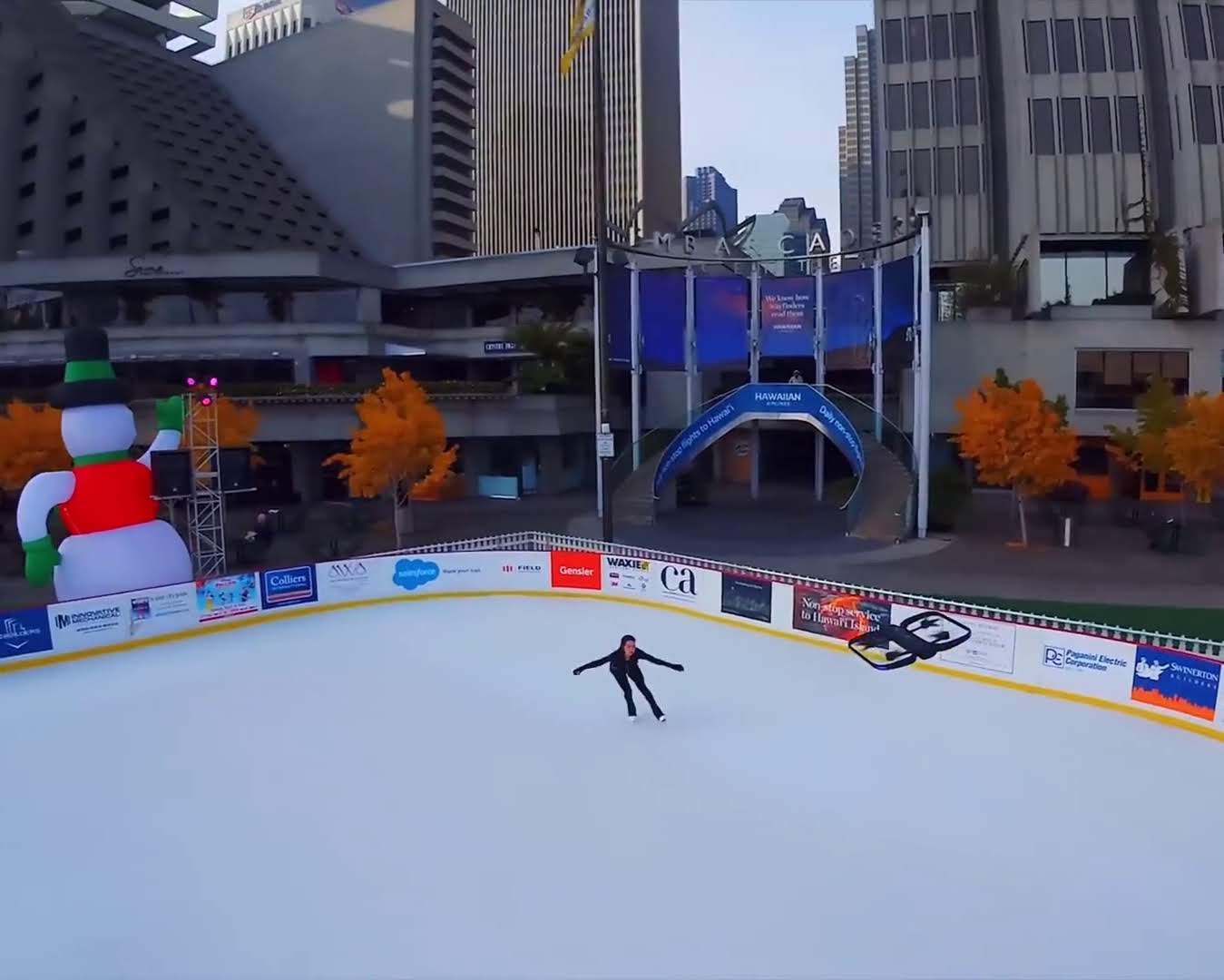} 
        \caption{{  A UAV is filming an athlete from various viewpoints~\citep{Skydio}}}
    \end{subfigure}
    \begin{subfigure}[t]{0.205\textwidth}
       \includegraphics[width=\linewidth]{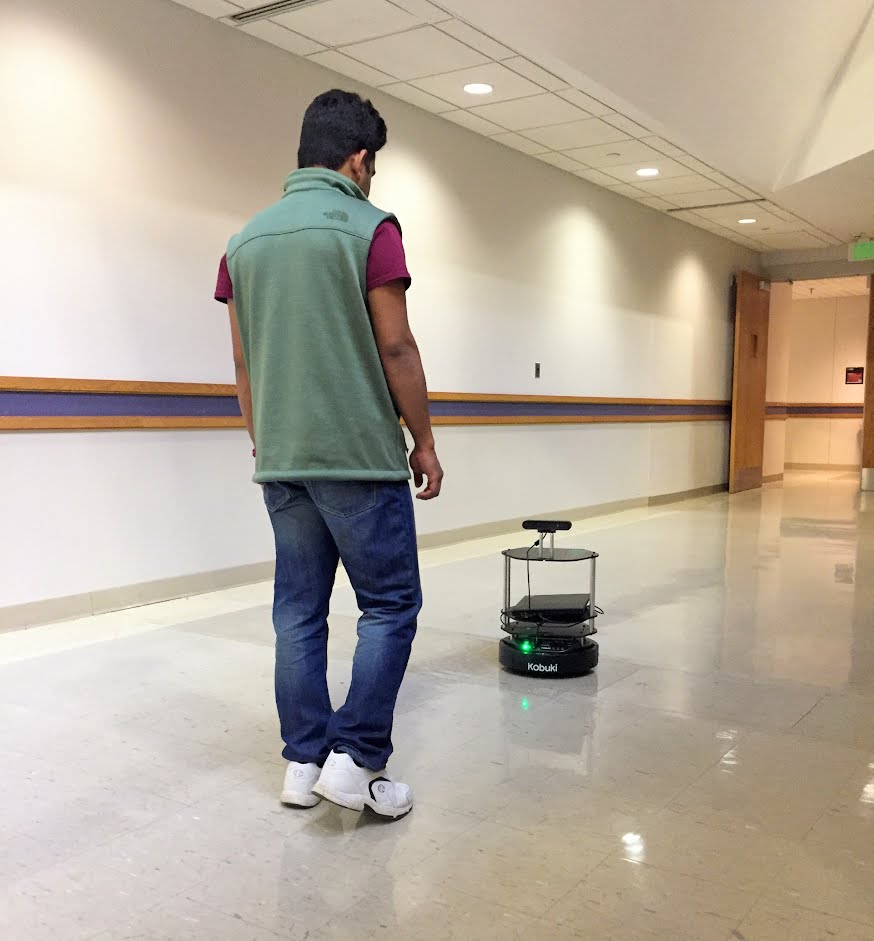}
        \caption{A UGV is leading a person through a hallway}
    \end{subfigure}
    \vspace{-2mm}
    \caption{Scenarios where a robot is not following its companion from behind.}
    \label{fig:inter5}
    \end{figure}

In recent years, researchers have begun to explore the particularities of different scenarios (Figure~\ref{fig:inter5}) where the robot should be in front or on the side of the person while following~\citep{hu2014design,ferrer2013robot,nagyflying}. These scenarios impart more operational challenges since the robot needs to predict the motion trajectory of the person, and needs some way to recover from a wrong prediction or action. Motion history and gaze behaviors of the person, and prior knowledge about the environment or destination can be utilized to model such anticipative behaviors. The person can help the robot make decisions in critical situations as well (using hand gestures or voice commands). Nevertheless, these aspects demand more research attention and experimental evaluations in real-world settings.

\subsection{Learning to Follow from Demonstration}
End-to-end learning of autonomous robot behaviors from demonstration is an interesting ongoing research topic. Researchers have reported exciting results in the domains of 2D robot navigation in cluttered environments~\citep{pfeiffer2017perception}, simple autonomous driving~\citep{codevilla2018end}, imitating driving styles~\citep{kuderer2015learning}, etc.
These results indicate that the end-to-end learning models, particularly the idea of \textit{learning from demonstration} can be very effective for person-following robots. Further research attention is required to explore other end-to-end (deep) learning-based models as well because they have the potential to significantly simplify autonomous person-following. There are a few research efforts already in this regard in simulation environments~\citep{pierre2018end,dewantara2016generation}; however, more extensive research and real-world experiments are necessary.

\subsection{Human-Robot Communication}
A generic communication paradigm for human-robot dialogue~\citep{thomason2015learning} can be very useful in practice for person-following applications. 
Several human-to-robot communication paradigms using speech, markers, and hand gestures are already discussed in this paper. There are not many research studies on how a robot can initiate communication and maintain a proper dialog with the human, particularly in applications where interactive user interfaces are not feasible~\citep{fulton2018robot}. Furthermore, effective and efficient ways of assessing risks in human-robot dialogue~\citep{robinette2016investigating,sattar2011towards} is another potential research problem in this domain.

\subsection{Enabling Social and Spatial Awareness}
Various forms of implicit human-robot interactions, particularly the preferred spatial and motion behaviors for person-following robots were discussed in the previous Section. Robots that are deployed in a social setting should be \textit{aware} of these aspects and the social norms in general~\citep{honig2018towards,kim2014social,granata2012framework}. 

\begin{figure}[b]
    \centering 
    \begin{subfigure}[t]{0.24 \textwidth}
        \includegraphics[width=\linewidth]{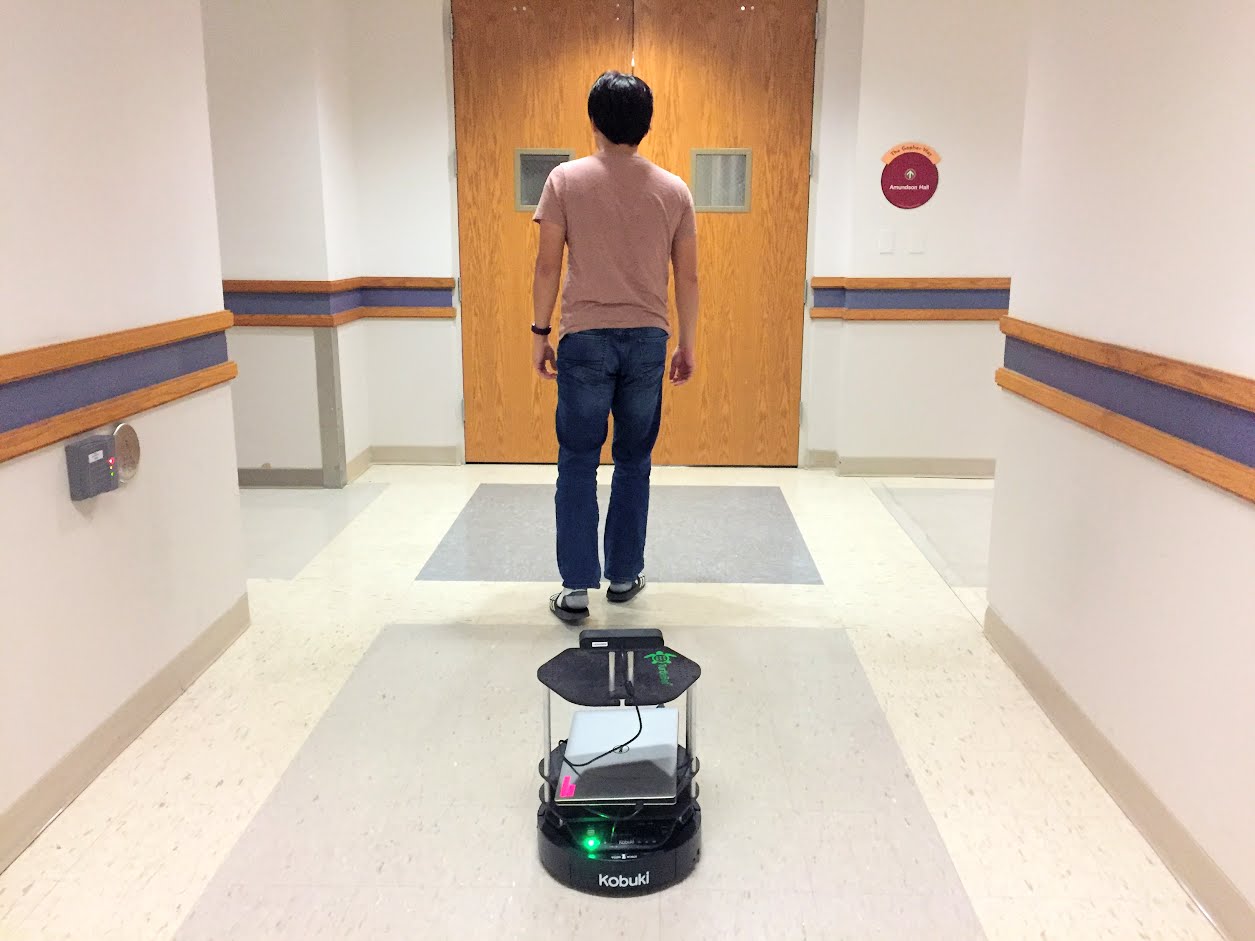} 
        \caption{A UGV is following a person while staying behind}
    \end{subfigure}
    \begin{subfigure}[t]{0.24\textwidth}
        \includegraphics[width=\linewidth]{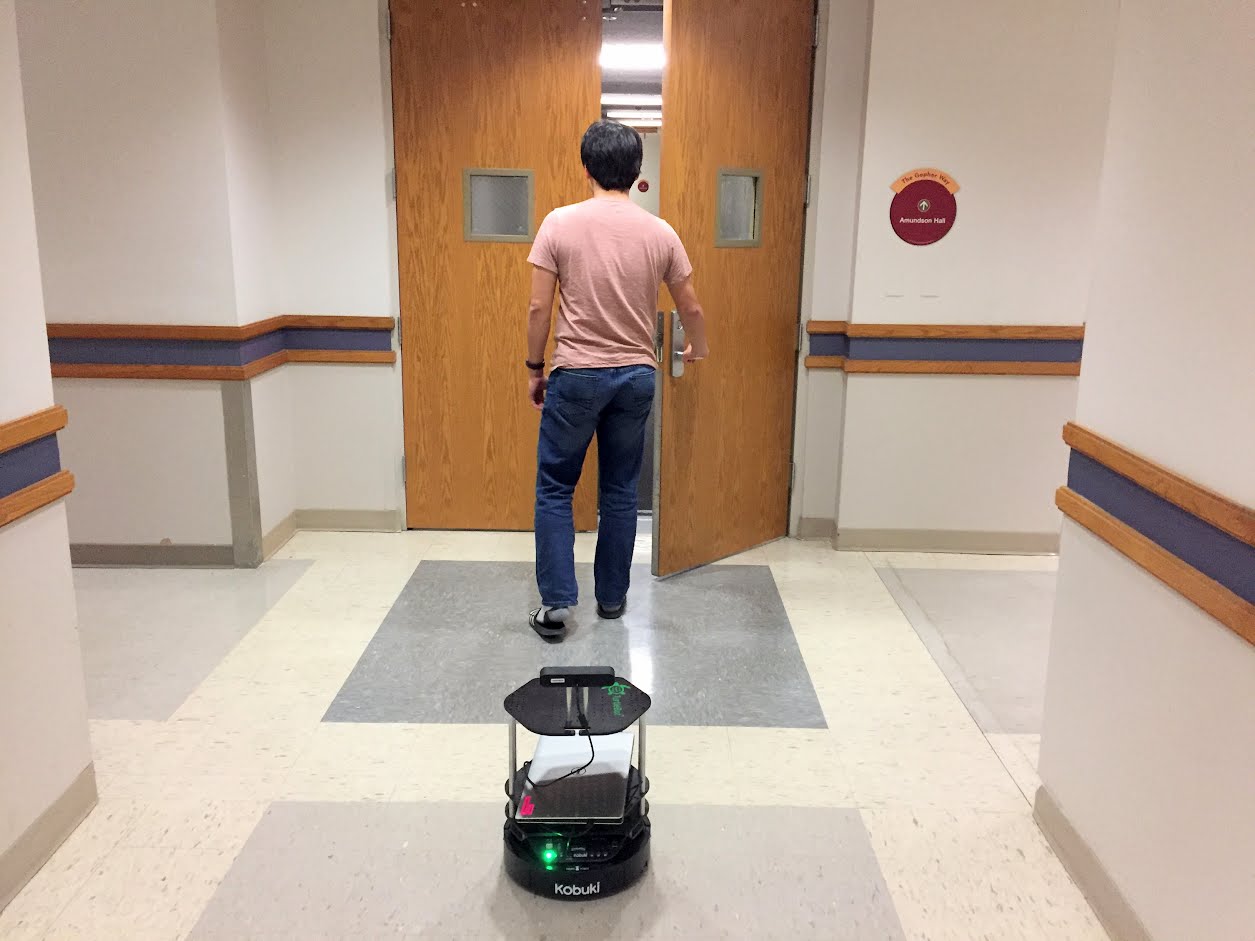}
        \caption{The UGV is standing clear of the door-opening}
    \end{subfigure}
    \vspace{-2mm}
    \caption{Illustration of a desired robot behavior: notice that the UGV is giving extra space to the person for opening the door.}
    \label{fig:inter6}
    \end{figure}

A particular instance of anticipative robot behavior is illustrated in Figure~\ref{fig:inter6}. Here, the robot anticipated the door-opening action~\citep{zender2007human}, increased the distance to the person by slowing down, and kept waiting instead of moving forward. Many other anticipated behaviors such as moving slowly while entering cross-paths, waiting at a side when the person is interacting with other people, etc., are important features of a social robot. These are difficult to quantify and implement in general~\citep{chen2017socially,kruse2013human}; extensive experiments and further user studies are required to model these social norms for person-following robots.

\subsection{Long-term Interaction and Support}
Another social aspect of the person-following UGVs is long-term interaction with their human companions. It has numerous potential applications in healthcare; for instance,~\cite{coninx2016towards} showed that long-term child-robot interaction was useful for learning and therapeutic purposes; ~\cite{chen2017effect},~\cite{kidd2008robots} proved that long-term interaction with a robot helped people in physical exercises. These, among many other studies, show that robots can help more by learning general behaviors and routine activities of their companion humans. Thorough analysis and user studies are needed to discover the feasibilities and utilities of long-term interactions for other person-following applications.

\subsection{Specific Person-Following}
Following a \textit{specific} person is generally more useful than following \textit{any} person, specially in a multi-human setting~\citep{satake2009robust} and in social or crowded environments. Moreover, the ability to follow a specific person is an essential feature for UGVs that accompany the elderly and people with disabilities~\citep{ilias2014nurse,liem2008hybrid}. It is straightforward to achieve this in some applications, with the use of an additional human face or body-pose recognition module~\citep{yoshimi2006development,cao2013specific}. However, scenarios such as following a person in crowded surrounding~\citep{germa2009vision}, avoiding an impeding person~\citep{hoeller2007accompanying}, etc., are rather challenging. Furthermore, lack of person-specific features while viewing a diver from behind (multiple divers may wear similar suits), make it a harder problem for underwater robots~\citep{xia2018visual}. Detecting a specific person from a distant UAV is also challenging due to similar reasons.

\subsection{Person Re-identification}
Several mechanisms for person \textit{recovery} or \textit{re-identification} used by existing person-following systems were mentioned in Section~\ref{reid_p}.   
They mostly use feature-based template matching ~\citep{koide2016identification,do2017reliable,gupta2017novel} techniques; trajectory replication-based techniques~\citep{chen2017integrating} are also used for re-identification when the target person transiently disappears from the robot's view and appears again. 
A number of recently proposed appearance-based deep models~\citep{ahmed2015improved,li2014deepreid} have significantly improved the state-of-the-art performance for person re-identification on standard datasets.    
Despite the potentials, these models are yet to be used in person-following systems. Investigating the applicability of these person re-identification models for specific person-following in human-dominated social settings$-$ is an interesting and potentially rewarding research direction.

\subsection{Surveillance and Rescue Support} 
Features such as person re-identification and adversarial person-following are useful for autonomous human surveillance using UAVs~\citep{portmann2014people}. Additionally, human rescue missions using a team of UAVs are invaluable in adversarial conditions~\citep{doherty2007uav}. These are critical applications and there are always rooms for further improvements.

\subsection{Embedded Parallel Computing Solutions}
As mentioned earlier, deep learning-based models provide robust solutions to most of the perception problems involved in person-following scenarios. One practical limitation of these models is that they are often computationally expensive and require parallel computing platforms. Therefore, faster mobile supercomputers and embedded parallel computing solutions~\citep{Jetson,coral} are going to be immensely useful in person-following applications. 
The recent success of the person-following UAV named Skydio R1~\citep{Skydio} is a practical example of that. However, high power consumption of these on-board computers is still a major concern; for instance, flight-time for a Skydio R1 is only about $16$ minutes. In addition to computational capacity and power consumption, many other aspects of mobile supercomputers such as durability, cooling mechanisms, etc., require further technological improvements. Future advancements in ultra-low power computer vision (TinyVision) and machine learning (TinyML) techniques and platforms~\citep{tinyml} can potentially play an important role in this regard.

\subsection{Addressing Privacy and Safety Concerns}
There has been an increasing amount of concerns across cyberspace about the privacy and safety issues of autonomous robots, particularly UAVs operating in social and public environments~\citep{privUAV}. A recent study~\citep{pewUAV} has found that about $54$\% of the US population thinks that drones and autonomous UAVs should not be allowed to fly near people’s homes. This is due to the fact that it undermines people's ability to assess the context and measure trust. While person-following UAVs are mostly used for recreational purposes in public areas and often crowded places, these concerns need to be addressed using technological and educational solutions~\citep{finn2012unmanned,wang2016flying} in order to ensure transparency and trust.

\section{Conclusion}\label{sec:con}
{  
Person-following by autonomous robots have numerous important applications in industry. Their usage in social settings and for entertainment purposes have flourished over the last decade as well. Researchers have approached various aspects of the \textit{autonomous person-following} problem from different perspectives and contributed to the development of a vast literature. 
This paper makes an effort to present a comprehensive overview of this large body of literature in a categorical fashion.
First, the design issues and operational challenges for person-following robots in ground, underwater, and aerial scenarios are presented. 
Then, the state-of-the-art methods for perception, planning, control, and interaction of various person-following systems are elaborately discussed. 
{
In addition, several operational considerations for applying these methods, underlying assumptions, and their feasibility in different use-cases are analyzed and compared in qualitative terms. }
Finally, a number of open problems and potential applications are highlighted for future research; improved solutions to these problems will significantly strengthen the literature and bridge the gap between research and practice. 
}

%\bibliographystyle{SageV}
%added comma in line 230 for 'et at.,' instead of 'et al.' 

\bibliographystyle{SageH}
\bibliography{ijrr_main}

%%%%%%%%%%%%%%%%%%%%%%%%%%%%%%%
%\begin{acks}
%%TODO
%\end{acks}
%%%%%%%%%%%%%%%%%%%%%%%%%%%%%%%

\end{document}